\documentclass[journal]{IEEEtran}
\usepackage{amsmath,amsfonts,amssymb}
\usepackage{algorithmic}
\usepackage{algorithm}
\usepackage{array}
\usepackage{booktabs}
\usepackage{multirow}
\usepackage[caption=false,font=normalsize,labelfont=sf,textfont=sf]{subfig}
\usepackage{textcomp}
\usepackage{stfloats}
\usepackage{url}
\usepackage{verbatim}
\usepackage{graphicx}
\usepackage{cite}
\usepackage{enumitem}
\newsavebox{\sigtablebox}
\begin{document}
\bstctlcite{IEEEexample:BSTcontrol}
\raggedbottom
\sloppy

\title{Zipper-LoRA: Dynamic Parameter Decoupling for Speech-LLM based Multilingual Speech Recognition}


\author{Yuxiang Mei, Delai Qiu, Shengping Liu, Jiaen Liang, 
Yanhua Long,~\IEEEmembership{Member,~IEEE}
\thanks{Corresponding author: Yanhua Long.  
Yuxiang Mei, Yanhua Long are with the Shanghai Engineering Research Center of Intelligent Education and Bigdata, Shanghai Normal University, Shanghai, 200234, China. Yanhua Long is also with the SHNU-Unisound Natural Human-Computer Interaction Lab, Shanghai Normal University. (e-mail: m153517@icloud.com; yanhua@shnu.edu.cn). 
Delai Qiu, Shengping Liu, Jiaen Liang are with the Unisound AI Technology Co., Ltd., Beijing, China (e-mail: qiudelai@unisound.com; liushengping@unisound.com; liangjiaen@unisound.com).
This work was done while Yuxiang Mei was an intern at Unisound.}}



\maketitle

\begin{abstract}

Speech Large Language Models (Speech-LLMs) have emerged as a powerful approach for 
automatic speech recognition (ASR) by aligning speech encoders with large language models. 
However, adapting these systems to multilingual settings with imbalanced data distributions remains 
challenging. In such scenarios, a stability-plasticity dilemma often arises: fully shared Parameter-Efficient 
Fine-Tuning (PEFT) can cause negative inter-lingual interference for under-represented languages, 
while fully language-specific tuning limits the cross-lingual beneficial knowledge transfer 
needed for low-resource tasks. To address this, we propose \textit{Zipper-LoRA}, 
a novel rank-level decoupling framework with three variants (Static, Hard, and Soft) 
that dynamically synthesizes LoRA updates from shared and language-specific subspaces.
By using a \textit{lightweight language-conditioned router}, 
Zipper-LoRA dynamically controls the contribution of each subspace at the LoRA rank-level, 
enabling fine-grained sharing where languages are compatible and strict decoupling when conflicts occur.
To further stabilize optimization under imbalanced data, we propose a two-stage training strategy with 
an \textit{Initial-B warm-start} that significantly accelerates convergence. 
Experiments on a 12-language mixed-resource setting show that Zipper-LoRA consistently 
outperforms both fully shared and independent baselines, 
particularly in extremely low-resource scenarios. Moreover, we demonstrate that these gains 
are robust across both chunked and non-chunked encoder configurations, confirming the framework’s 
reliability for practical, large-scale multilingual ASR. 
Our code and data will be available at \url{https://github.com/YuCeong-May/Zipper-LoRA} for reproducibility.

\end{abstract}


\section{Introduction}

\IEEEPARstart{T}{he} rapid progress of Large Language Models (LLMs) has fundamentally 
reshaped the landscape of modern artificial intelligence. Recent foundation models 
\cite{singh2025openai,arxiv2026llama4herdarchitecture,yang2025qwen3} demonstrate emergent reasoning and 
generation capabilities, motivating a paradigm shift towards architectures that bridge powerful 
speech encoders with LLM backbones via projection interfaces 
\cite{an2025fun,bai2024seed,xu2025fireredasr,tang2023salmonn,chen2025minmo}. 
These unified Speech-LLM systems \cite{wang25m_interspeech,xu2025qwen2,hu25f_interspeech} not only 
achieve high-performance Automatic Speech Recognition (ASR) but also enable complex instruction-following 
behaviors for speech-centric reasoning tasks.

Despite these advancements, robust and fast domain adaptation of a higher-resource well-trained 
Speech-LLM model to low-resource or resource-imbalanced multilingual ASR scenarios remains challenging. 
Current Speech-LLMs are predominantly optimized for high-resource languages (e.g., English, Mandarin), 
when extended to a long-tailed data distribution that contains numerous low-resource languages, 
ASR performance often degrades significantly \cite{multilingual_degrade}. 
A primary bottleneck is inter-lingual interference: gradient updates dominated by high-resource training 
data may distort the shared representations required for under-represented languages. 
This issue essentially reflects the classic stability-plasticity dilemma in continual learning 
and multi-task adaptation \cite{stability_plasticity}: an ideal model needs to stay stable 
enough to handle shared acoustic features while remaining flexible enough to pick up 
on specific phonological and acoustic differences between languages.

For an ASR adaptation task, directly fine-tuning billion-parameter Speech-LLMs is unrealistic 
due to the massive computation and memory required. This is why Parameter-Efficient Fine-Tuning (PEFT) 
has become the standard, with Low-Rank Adaptation (LoRA) \cite{lora} being the most common choice. 
However, using PEFT in diverse multilingual environments brings out a fundamental conflict between 
reducing inter-lingual interference and encouraging cross-lingual positive knowledge transfer. 
A typical design shares a single LoRA module across all languages \cite{universal_lora}, 
but this fully shared parameterization entangles optimization trajectories; when training data is imbalanced, 
high-resource languages dominate the shared space and cause negative knowledge transfer \cite{negative_transfer}.
On the other hand, language-specific LoRA assigns an independent module per language \cite{zheng2026language}, 
which eliminates inter-lingual interference but blocks cross-lingual positive transfer.
Low-resource languages, lacking sufficient supervision, fail to benefit from universal 
acoustic and linguistic features (e.g., shared phonemes or articulatory patterns) 
learned from data-rich languages, resulting in suboptimal generalization.

Recent research works have explored dynamic and modular LoRA variants to bridge this gap. 
For example, the proposed Mixture-of-Experts (MoE) adaptations \cite{li25p_interspeech,lu2025hipa,li2024mixlora} 
attempt to scale model capacity via multiple experts and routing mechanisms. 
However, most MoE approaches operate at a coarse granularity (e.g., selecting entire adapter modules or layers) 
and typically rely on token-level routing that does not explicitly guarantee the decoupling of linguistic attributes. 
Other methods, such as FlyLoRA \cite{flylora2025} or DyLoRA \cite{dylora}, focus on adjusting 
effective rank for compression or efficiency, yet they lack a mechanism to explicitly distribute 
capacity between shared and specific components based on language identity. 
The key research challenge lies in designing a fine-grained mechanism that 
maximizes the sharing of transferable cross-lingual acoustic knowledge while strictly 
isolating harmful inter-lingual interference.

To address this challenge, we propose \textbf{Zipper-LoRA}, a novel rank-level dynamic decoupling 
framework for multilingual Speech-LLM system adaptation. Drawing inspiration from the mechanical action of a zipper, 
our method dynamically synthesizes the LoRA adaptation matrix by interlocking two complementary subspaces: 
1) a \textit{Shared Subspace} that captures universal acoustic and linguistic regularities, and 
2) a \textit{Specific Subspace} that models language-specific phonological patterns and acoustic attributes. 
Unlike static MoE, Zipper-LoRA employs a lightweight router conditioned on language identity 
to control the contribution of these components at the fine-grained rank-level. 
By dynamically `zipping' together shared features and `unzipping' conflicting ones at the rank level, 
Zipper-LoRA provides a flexible solution to balance stability and adaptability. 
This fine-grained control allows the model to pool cross-lingual knowledge where languages align 
while blocking inter-lingual interference where they differ.

We evaluate Zipper-LoRA on a standard, high-resource well-trained Speech-LLM architecture that 
connects a speech encoder, a trainable projector and a text LLM, and adapt it to a 12-language 
mixed-resource multilingual ASR setting. Our extensive results demonstrate that Zipper-LoRA delivers 
substantial performance gains. It improves significantly over fully shared LoRA 
by alleviating inter-lingual interference, and compared with fully independent-LoRA, 
it achieves better performance while using fewer parameters, confirming its ability to unlock positive 
transfer with higher efficiency. Our primary contributions are summarized as follows:
\begin{itemize}[leftmargin=0pt,itemindent=2.5em]
  \item \textbf{Dynamic Rank-Level LoRA Decoupling Framework}: 
  We propose Zipper-LoRA, a novel PEFT architecture that dynamically synthesizes LoRA 
  adaptation matrices by ``zipping" together shared and language-specific subspaces at the rank level. 
  We introduce three distinct variants: Zipper-LoRA-Static (hard partitioning), Zipper-LoRA-Hard (binary column selection), and Zipper-LoRA-Soft (dynamic rank-wise mixing), to provide a flexible range of solutions. 
  This fine-grained composition allows the model to better balance cross-lingual knowledge sharing 
  with the isolation of inter-lingual interference, significantly outperforming coarse-grained MoE approaches.

  \item \textbf{LID-Aware Contextual Routing}: 
  We propose a lightweight, rank-level routing mechanism powered by Whisper-derived 
  language identity (LID) embeddings. Unlike traditional MoE methods that switch between 
  entire modules, our router performs fine-grained composition by dynamically controlling 
  the contribution of shared and language-specific subspaces at the individual rank level, 
  ensuring optimal knowledge transfer tailored to each language's needs.

  \item \textbf{Two-Stage Training with Initial-B Warm-start}:
  To stabilize optimization under imbalanced multilingual data, we develop a robust training strategy. 
  By decoupling cross-modal alignment from language-specific adaptation and employing an Initial-B warm-start, 
  which initializes the low-rank up-projection from a converged Zipper-LoRA model, we significantly 
  accelerate convergence and ensure more stable learning of both shared and specific subspaces.

  \item \textbf{Robustness for Practical Deployment}: 
  We verify that Zipper-LoRA’s performance gains are consistent across diverse encoder configurations, 
  including both chunked and non-chunked processing. By demonstrating that the framework remains 
  effective regardless of input processing constraints, we confirm its reliability for practical, 
  large-scale multilingual ASR applications.

\end{itemize}

\section{Related Work}
\label{sec:related}

\subsection{Large Language Models for Speech Processing}
\label{subsec:llm}

Speech processing has increasingly shifted from task-specific pipelines toward unified Speech Large 
Language Models (Speech-LLMs), where a speech/audio encoder interfaces with a text LLM through a 
lightweight alignment module (e.g., linear/MLP projection or a small cross-modal adapter). Earlier 
systems often followed a cascade paradigm that combines an ASR front-end with a text-only LLM for 
downstream reasoning, while recent end-to-end Speech-LLMs integrate speech perception and language 
generation more tightly and support instruction-following behaviors beyond speech recognition 
\cite{speechgpt2023,audiopalm2023,salm2023,tang2023salmonn}.
A common design builds upon strong pretrained speech encoders trained on large-scale multilingual 
speech-text data (e.g., Whisper) \cite{whisper}, and connects them to an LLM backbone via a 
lightweight interface for representation alignment and conditioning \cite{blip2_2023,llava_2023}.
Along this line, Speech-LLM systems such as SALM \cite{salm2023} and SALMONN \cite{tang2023salmonn} 
couple audio encoders with (often frozen) LLM backbones and introduce lightweight adaptation 
modules (e.g., LoRA) to enable ASR and broader audio-language understanding.
Representative audio-language foundation models include Qwen-Audio \cite{qwenaudio2023} and 
speech-centric variants such as WavLLM \cite{wavllm}, which further demonstrate the feasibility of 
unifying speech perception with LLM-style generation.
Despite strong general capabilities, these models are often biased toward high-resource languages 
that dominate the pretraining distribution, and adapting Speech-LLMs to multilingual settings with 
long-tailed language distributions remains challenging, especially for low-resource languages with 
limited supervision \cite{googleusm2023,pratap2023mms,seamlessm4t2025,omnilingual2025omnilingual}.

\subsection{Parameter-Efficient Fine-Tuning in Multilingual ASR}
\label{subsec:peft}

Fine-tuning billion-parameter models is computationally expensive and can be unstable in 
data-limited regimes. Parameter-efficient fine-tuning (PEFT) therefore plays a central role in adapting 
large models to specific domains and languages 
\cite{peftsurvey2024,houlsby2019adapters,prefixtuning2021,prompttuning2021}.
Among PEFT methods, low-rank adaptation (LoRA) has become a widely adopted choice due to its 
simplicity, training stability, and minimal inference overhead \cite{lora}. Recent variants 
further improve practicality and performance, including activation-scaling approaches such as 
IA$^{3}$ \cite{llava_2023}, quantized fine-tuning such as QLoRA \cite{qlora2023}, and adaptive/weight-
decomposed variants such as AdaLoRA and DoRA \cite{adalora2023,dora2024}.
In multilingual ASR, LoRA-style adaptation exposes a recurring tension between cross-lingual 
sharing and language-specific specialization. A shared adaptation module encourages 
collaborative acoustic and linguistic knowledge transfer
by reusing parameters across languages; however, it remains vulnerable to 
negative interference under imbalanced multilingual training, where updates dominated by high-
resource languages can suppress learning signals for low-resource languages 
\cite{wu2020negativeinterference}.
Conversely, language-conditioned or language-specific adaptation reduces such 
interference but may limit cross-lingual positive transfer and increase the parameter footprint 
with the number of languages \cite{chen2023languageprompt,li25p_interspeech,lora_whisper2024,zhao2025samd,zhao2026lors}.
These observations motivate designs that explicitly allocate capacity to both shared and 
language-specific components, aiming to retain cross-lingual knowledge transfer while 
controlling interference in long-tailed multilingual ASR.

\subsection{Dynamic LoRA Variants and Mixture-of-Experts}
\label{subsec:dlora}

To increase the expressivity of parameter-efficient adaptation, recent work has explored dynamic 
and modular designs. Dynamic-rank approaches such as DyLoRA\cite{dylora} train LoRA blocks that can be truncated 
to different effective ranks without expensive search, enabling flexible rank selection. Beyond rank dynamics, Mixture-of-Experts (MoE) style LoRA introduces multiple 
expert updates and uses routing mechanisms to select or combine experts for each input 
\cite{mole2024,loramixer2025,ase2025}. While effective for scaling adaptation capacity, 
many MoE-LoRA variants operate at relatively coarse granularity (e.g., selecting experts at 
layer/module level), and routing can suffer from imbalance or collapse without careful 
regularization \cite{switchtransformers2021}.
Parallel lines of work explore alternative parameterizations of low-rank updates, such as FourierFT 
that learns a small set of spectral coefficients to reconstruct weight updates efficiently 
\cite{fourierft2024}, and rank-wise MoE-inspired designs such as FlyLoRA that aim to mitigate 
interference and improve decoupling \cite{flylora2025}.
Overall, these methods primarily target efficiency and general expressivity, but they are typically 
language-agnostic and do not explicitly enforce a boundary between universal acoustic knowledge and 
language-specific characteristics.
In contrast, our work targets multilingual decoupling in LLM-based speech recognition with explicit 
language awareness and fine-grained control: Zipper-LoRA performs rank-level dynamic routing to 
allocate LoRA capacity between shared and language-specific subspaces conditioned on language 
identity, enabling selective sharing of transferable acoustic regularities while isolating 
language-specific interference.

\label{subsec:spllmarch}
\begin{figure}[t]
  \centering
  \includegraphics[width=\linewidth]{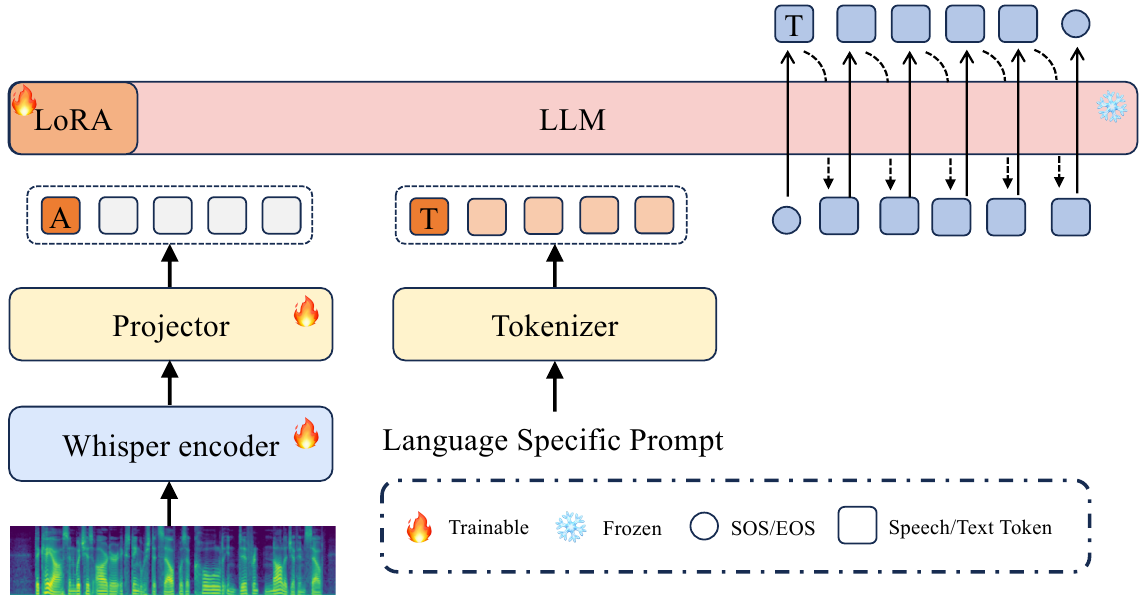}
  \caption{Overall Speech-LLM backbone consisting of a speech encoder, a modality projector, and a decoder-only LLM.}
  \label{fig:overall_arch}
\end{figure}

\section{Foundations of Speech-LLM and Adaptation Paradigms}
\label{sec:foundations}

In this section, we establish the technical foundations for multilingual Speech-LLM adaptation. We 
begin by defining the backbone architecture and the language-specific prompting mechanism. Then, 
we provide a comprehensive overview of representative Parameter-Efficient Fine-Tuning (PEFT) paradigms, 
ranging from foundational LoRA structures to recent rank-wise Mixture-of-Experts (MoE) adaptations.

\subsection{Speech-LLM Architecture}

\begin{figure}
  \centering
  \includegraphics[width=0.85\linewidth]{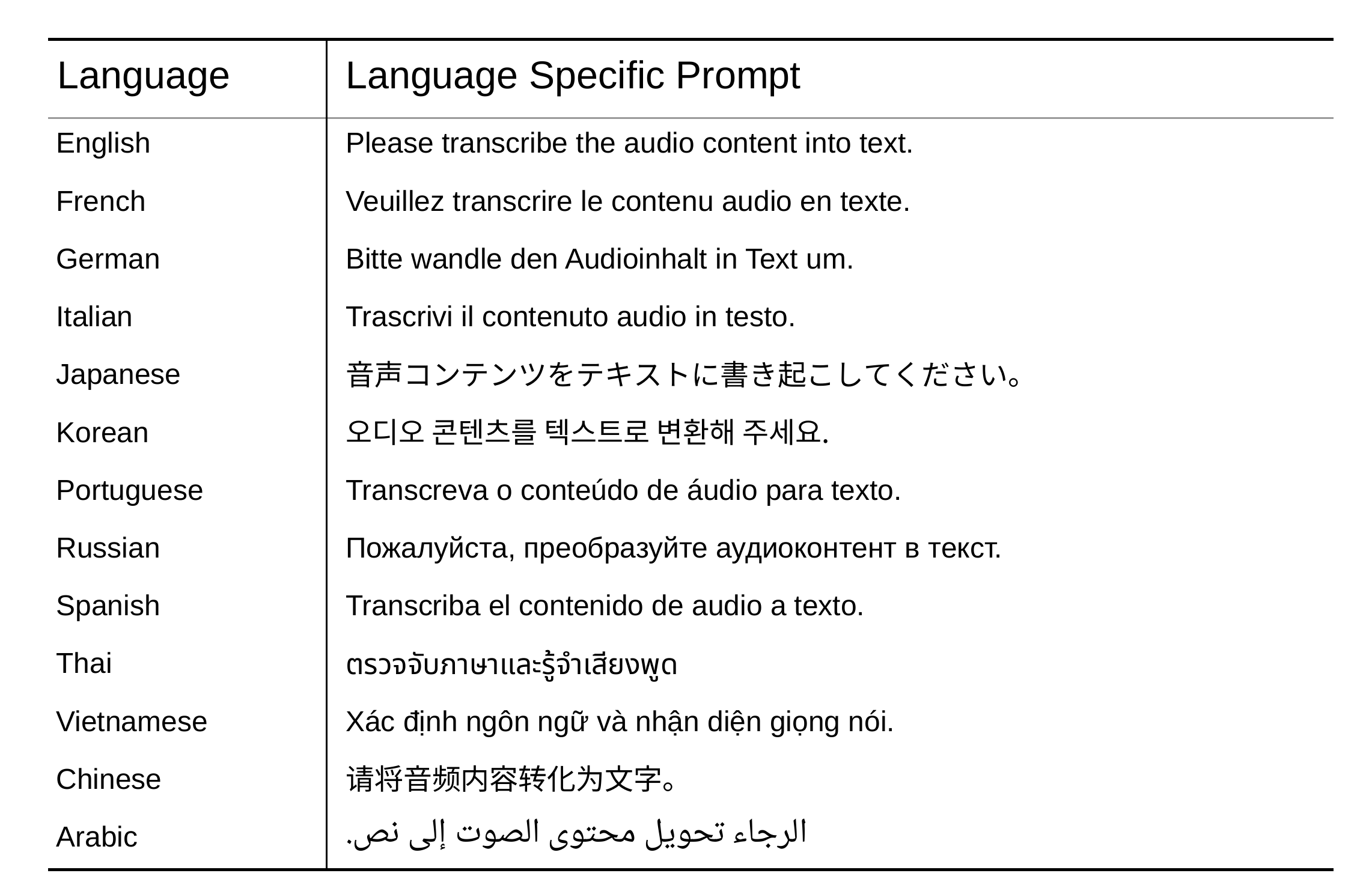}
  \caption{Language-specific prompts. All these prompts have the same meaning: ``Please transcribe the audio content into text." but are written in specific languages based on the language given for a speech.}
  \label{fig:prompt}
\end{figure}

\begin{figure*}[ht]
  \centering
  \includegraphics[width=0.85\linewidth]{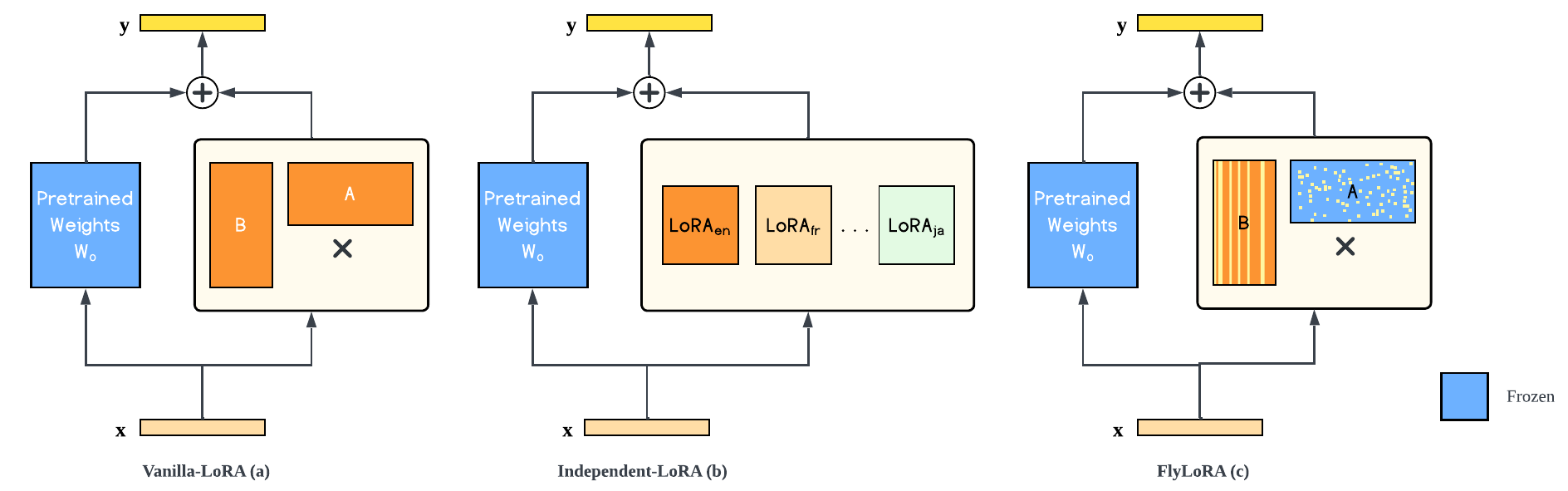}
  \caption{Illustration of three representative PEFT frameworks for multilingual ASR Speech-LLM adaptation: Vanilla-LoRA (a), Independent-LoRA (b), and FlyLoRA (c).}
  \label{fig:lora}
\end{figure*}

Our system adopts a unified Speech-LLM paradigm. As illustrated in Fig.~\ref{fig:overall_arch}, the 
backbone consists of three primary components: a speech encoder $\mathcal{E}$, a modality projector 
$\mathcal{P}$, and a large language model $\mathcal{M}$. The process begins with the general 
formulation of audio-to-text transformation. 
Given an input speech sequence $\mathbf{X}$, the speech encoder first extracts high-level acoustic 
representations:
\begin{equation}
  \mathbf{H}_{enc} = \mathcal{E}(\mathbf{X})
\end{equation}

To ensure these features are interpretable by the language model, the modality projector $\mathcal{P}$
maps them into the LLM’s token embedding space:
\begin{equation}
  \mathbf{H}_{proj} = \mathcal{P}(\mathbf{H}_{enc})
\end{equation}
In a standard setting, the decoder-only LLM $\mathcal{M}$ directly takes 
these projected embeddings as input to produce the target text sequence $\mathbf{Y}$ 
through $\mathbf{Y} = \mathcal{M}(\mathbf{H}_{proj})$, optimized via the standard 
cross-entropy loss.

However, to effectively handle multilingual ASR discrimination, we enhance this standard pipeline by 
introducing a \textit{Language-Specific Prompting} mechanism. As shown in Fig.~\ref{fig:prompt}, we 
construct a set of prompts that share the same semantic meaning, namely “Please transcribe the audio 
content into text.” These prompts are expressed in different languages according to the target speech 
language. By providing these explicit linguistic cues, we enable the LLM to generate transcriptions in 
the intended language, thereby reducing language-switching errors. Formally, the prompt is introduced in the text embedding space before the projected speech 
representations. Let $\mathrm{Emb}(\cdot)$ denote the LLM token embedding lookup and 
$\oplus$ denote sequence concatenation. For a target language $l$, we first construct the prompt token 
sequence $\mathbf{T}_{prompt}^{(l)}$ and obtain its embedding representation:
\begin{equation}
\mathbf{E}_{prompt}^{(l)} = \mathrm{Emb}\!\left(\mathbf{T}_{prompt}^{(l)}\right)
\end{equation}
The prompt embeddings are then concatenated with the projected speech representations $\mathbf{H}_{proj}$ to form the final input sequence to the LLM:
\begin{equation}
\mathbf{Z}^{(l)} = \mathbf{E}_{prompt}^{(l)} \oplus \mathbf{H}_{proj}.
\end{equation}
Conditioned on this combined representation, the decoder-only LLM $\mathcal{M}$ generates the target transcription as,
\begin{equation}
\mathbf{Y} = \mathcal{M}(\mathbf{Z}^{(l)})
\end{equation}

\subsection{Representative PEFT Adaptations of Speech-LLM}
\label{subsec:raspeechllm}

To adapt the large-scale Speech-LLM architecture detailed in Sec.~\ref{subsec:spllmarch} for 
multilingual ASR, full parameter fine-tuning is often computationally expensive and memory intensive. 
Parameter-efficient fine-tuning (PEFT) methods therefore provide a practical alternative by updating 
only a small set of additional parameters while keeping the pretrained backbone frozen. Among various 
PEFT techniques, LoRA-based adaptation has emerged as one of the most widely used approaches for both 
LLMs and multimodal systems. Building upon this framework, we employ a dual-side LoRA strategy, 
applying LoRA variants to the speech encoder and standard LoRA on the LLM, to enable efficient 
multilingual adaptation. As shown in Fig.~\ref{fig:lora},  we review three 
representative PEFT frameworks, ranging from foundational approaches to recent advancements, which 
serve as strong baselines and the direct motivation for our proposed method.

Technically, for a frozen weight matrix $W_0 \in \mathbb{R}^{d_{out} \times d_{in}}$, LoRA adapts it by introducing a low-rank update:
\begin{equation}
W = W_0 + \Delta W
\end{equation}
where the layer output is computed as $W_0\mathbf{x} + \Delta W\mathbf{x}$. Different PEFT variants mainly differ in how $\Delta W$ is parameterized and shared across languages.

\textbf{Vanilla-LoRA}: As the most foundational approach, Vanilla-LoRA applies a single low-rank 
adapter that is fully shared across all languages. As shown in Fig.~\ref{fig:lora}(a), 
the LoRA branch parameterizes a rank-$r$ update as:
\begin{equation}
  \Delta W = \frac{\alpha}{r}BA
\end{equation}
where $A \in \mathbb{R}^{r \times d_{in}}$ and $B \in \mathbb{R}^{d_{out} \times r}$ are 
shared low-rank factors (down-/up-projection), and $\alpha$ is a scaling hyper-parameter.
Since both $A$ and $B$ are fully shared across languages, the same low-rank subspace is used to adapt $W_0$. 
While this promotes maximum cross-lingual knowledge transfer, it is highly susceptible to 
negative inter-lingual interference. Under imbalanced multilingual training, the shared 
subspace tends to be dominated by high-resource languages, which can suppress learning signals 
for low-resource languages \cite{wu2020negativeinterference}.

\textbf{Independent-LoRA}: To eliminate such inter-lingual interference, 
Independent-LoRA represents a traditional design choice that enforces strict parameter isolation. 
As illustrated in Fig.~\ref{fig:lora}(b),
while the pretrained $W_0$ remains shared and frozen, a dedicated LoRA module is assigned to each language $l \in \{1,\dots,L\}$:
\begin{equation}
\Delta W^{(l)} = \frac{\alpha}{r}B^{(l)}A^{(l)}
\end{equation}
This design effectively reduces inter-lingual interference by preventing parameter sharing in the LoRA subspace, 
but it also weakens positive knowledge transfer across languages, which can hurt low-resource languages.

\textbf{FlyLoRA}: A more recent advancement, FlyLoRA \cite{flylora2025}, introduces a rank-wise mixture-of-experts (MoE) 
mechanism into the adaptation process. As shown in Fig.~\ref{fig:lora}(c), it treats each rank component as an individual expert 
and dynamically activates a sparse subset of ranks via an implicit routing mechanism. 
Specifically, it employs a \emph{sparse and frozen} down-projection matrix $A$ to compute routing 
scores $\mathbf{y} = A\mathbf{x} + \mathbf{d}$, where $\mathbf{d} \in \mathbb{R}^{r}$ is a learnable bias term for load 
balancing. Based on these scores, a top-$k$ operator selects the active 
indices $\mathcal{I}_{\text{top}k} \subseteq \{1,\dots,r\}$. The adapter weight is then constructed by aggregating 
the top-$k$ active components:
\begin{equation}
  \Delta W = \frac{\alpha}{r} \sum_{i \in \mathcal{I}_{\text{top}k}} \mathbf{b}_i \mathbf{a}_i^\top
\end{equation}
where $\mathbf{b}_i$ and $\mathbf{a}_i^\top$ are the $i$-th column and row of the up- and down-projection matrices, 
respectively. Although FlyLoRA was originally proposed for general task decoupling in LLMs, 
its unique rank-wise MoE structure serves as a strong baseline and a direct inspiration for our work. 
However, since FlyLoRA relies on implicit routing without explicit language cues, it lacks a dedicated mechanism 
to decouple language-specific specialization from shared acoustic knowledge. 
This limitation motivates our proposed \textit{Zipper-LoRA}, which explicitly harmonizes these two components.

\section{Proposed Zipper-LoRA}
\label{sec:zipper_lora}

\begin{figure*}
  \centering
  \includegraphics[width=0.85\linewidth]{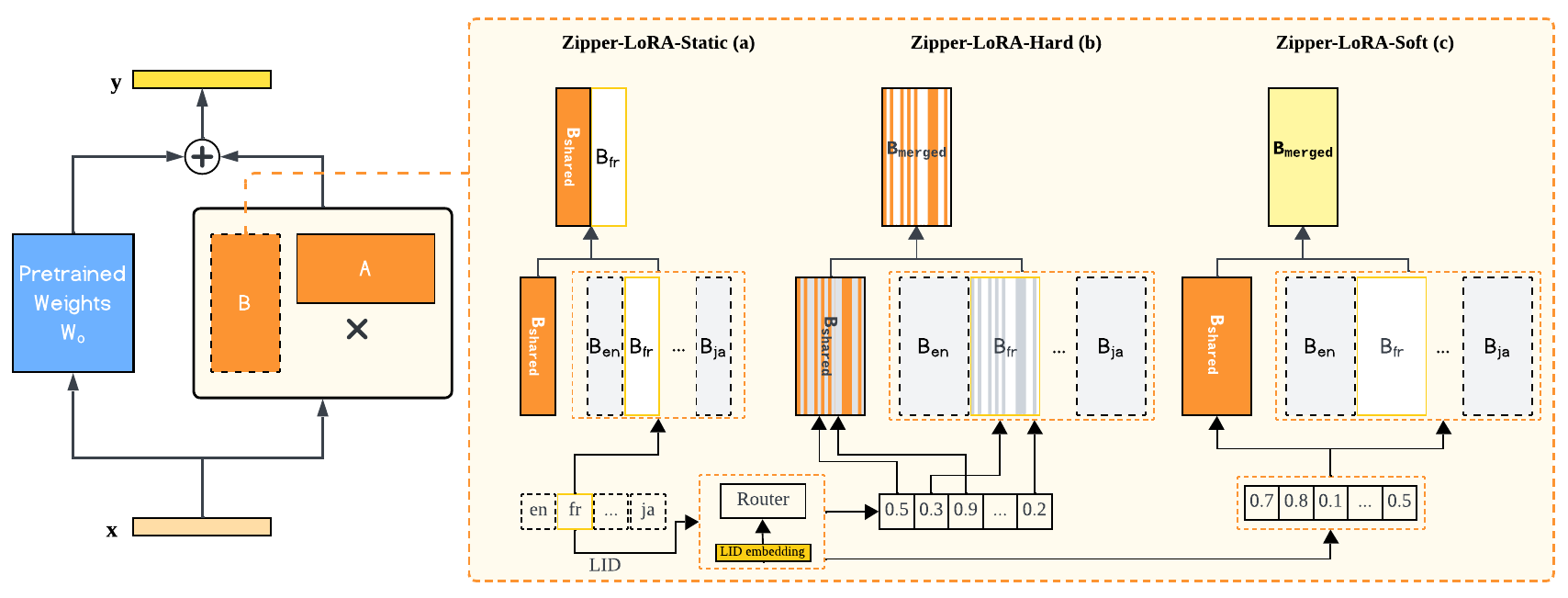}
  \caption{Overview of the proposed Zipper-LoRA. A language-aware router outputs rank-wise mixing weights from language embeddings to construct $B_{\text{merged}}^{(l)}$ for multilingual ASR adaptation.}
  \label{fig:zipper_lora}
\end{figure*}

As discussed in the above sections, the adaptation of multilingual ASR Speech-LLMs poses two competing 
requirements: 1) The model needs shared capacity to facilitate cross-lingual knowledge transfer, 
which is important for improving low-resource language performance; 2) It requires language-discriminative 
capacity to mitigate inter-lingual interference, particularly under the imbalanced multilingual data training. 
Throughout this process, the adaptation must remain parameter-efficient to stay scalable 
for large-scale frozen backbones, i.e., introducing only a small number of trainable parameters 
on top of a frozen backbone.

To harmonize these objectives, we propose \textbf{Zipper-LoRA}, a family of PEFT modules designed to explicitly 
allocate and integrate shared and language-specific rank components in a language-aware manner. As illustrated in 
Fig.~\ref{fig:zipper_lora}, all Zipper-LoRA variants leverage a globally shared down-projection matrix $A$, ensuring 
common acoustic-linguistic feature extraction. The fundamental distinction between these variants lies in 
the rank-wise construction of the up-projection matrix $B$.
Intuitively, Zipper-LoRA ``zips" together shared and language-specific columns of $B$ to form an effective rank-$r$ 
adapter for each language. This integration can be realized through either a static allocation or a dynamic rank-wise 
composition.

To provide a comprehensive understanding of the proposed framework, in the following sections, we first 
establish the definition of the shared down-projection matrix $A$ and the rank-wise construction of $B$. 
We then present the proposed Zipper-LoRA family, ranging from the static language-wise hard split (Zipper-LoRA-Static)
to dynamic binary selection (Zipper-LoRA-Hard) and soft mixing mechanisms (Zipper-LoRA-Soft). This is followed by the LID-aware contextual routing 
that manages these rank-wise compositions, concluding with the specialized training strategies 
tailored for optimizing Zipper-LoRA, such as segment-wise encoding and parameter initialization.

\subsection{Definition: Shared $A$ and Rank-wise Construction of $B$}
\label{subsec:definition}

The rank-level design provides a suitable granularity for multilingual PEFT. 
Each LoRA rank can be viewed as a compact adaptation direction, so composing shared and 
language-specific columns at the rank level allows the model to preserve transferable directions 
while isolating language-sensitive components. Compared with coarser layer-level or adapter-level 
routing, which switches or duplicates an entire module, rank-level composition offers finer 
sharing/decoupling granularity with little routing overhead. This is particularly useful for 
multilingual ASR, where languages may share some acoustic or phonetic factors while still requiring 
separation for conflicting language-specific patterns.

Let $A \in \mathbb{R}^{r \times d_{in}}$ be the shared down-projection matrix.
Zipper-LoRA constructs an effective up-projection matrix $B_{\text{merged}} \in \mathbb{R}^{d_{out}\times r}$ 
by combining a shared bank $B_{\text{shared}}$ and a language-specific bank $B_{\text{spec}}^{(l)}$.
The resulting adaptation is applied as
\begin{equation}
  \Delta W^{(l)}
  \;=\;
  \frac{\alpha}{r}\,B_{\text{merged}}^{(l)}\,A
  \label{eq:zipper_apply}
\end{equation}
where $\Delta W^{(l)}$ denotes the LoRA-induced weight update for language $l$, and $B_{\text{merged}}^{(l)}$ is instantiated differently by the three variants below.
In all cases, $B_{\text{merged}}$ is formed rank-wise, which matches the visualization in Fig.~\ref{fig:zipper_lora}(a)-(c).

\subsection{Zipper-LoRA-Static (Language-wise Hard Split)}
\label{subsec:zipperstatic}

Zipper-LoRA-Static implements a simple MoE-style specialization by hard partitioning 
rank components into shared and language-specific parts.
As shown in Fig.~\ref{fig:zipper_lora}(a), we assume the target rank is $r$ (e.g., $r{=}32$), 
and split it into $r_s$ shared ranks and $r_p$ language-specific ranks, with $r_s + r_p = r$ (e.g., $r_s{=}16$, $r_p{=}16$).
We learn a shared matrix $B_{\text{shared}} \in \mathbb{R}^{d_{out}\times r_s}$ and a language-specific matrix
$B_{\text{spec}}^{(l)} \in \mathbb{R}^{d_{out}\times r_p}$ for each language $l$.
The merged up-projection is formed by concatenation:
\begin{equation}
  B_{\text{merged}}^{(l)}
  \;=\;
  \big[\,B_{\text{shared}}, B_{\text{spec}}^{(l)}\,\big]
  \in \mathbb{R}^{d_{out}\times r}
  \label{eq:zipper_static}
\end{equation}
This design preserves a fixed amount of shared capacity while allocating a dedicated subspace for each language.

\subsection{Zipper-LoRA-Hard (Dynamic Binary Column Selection)}
\label{subsec:sipperlorahard}

Zipper-LoRA-Hard enables language-dependent specialization through \emph{binary} rank-wise 
selection between the shared bank and the language-specific bank.
As shown in Fig.~\ref{fig:zipper_lora}(b), we learn a shared bank
$B_{\text{shared}} \in \mathbb{R}^{d_{out}\times r}$ and a language-specific bank
$B_{\text{spec}}^{(l)} \in \mathbb{R}^{d_{out}\times r}$ (both of rank $r$).
Given a target language $l$, a language-aware router takes the corresponding language 
embedding $\mathbf{e}^{(l)}$ as input and outputs a rank-wise gating vector $\mathbf{p}^{(l)} \in [0,1]^r$.
Each element of this vector controls one LoRA rank column, so the routing decision is made at the 
rank level rather than at the adapter or layer level.
We then obtain a binary selection mask by thresholding:
\begin{equation}
  s_i^{(l)} = \mathbb{I} \left[p_i^{(l)} \ge \tau\right],\qquad
  \mathbf{s}^{(l)} \in \{0,1\}^r
  \label{eq:zipper_hard_mask}
\end{equation}
where $\tau$ is a fixed threshold. The merged up-projection is constructed by taking one column 
from either $B_{\text{spec}}^{(l)}$ or $B_{\text{shared}}$ for each rank position:
\begin{equation}
  B_{\text{merged}}^{(l)}
  \;=\;
  \mathrm{Zip} \Big(B_{\text{shared}},\, B_{\text{spec}}^{(l)},\, \mathbf{s}^{(l)}\Big)
  \label{eq:zipper_hard_zip}
\end{equation}
where $\mathrm{Zip}(\cdot)$ denotes the rank-wise ``zipper'' operation shown in Fig.~\ref{fig:zipper_lora}(b):
the $i$-th column of $B_{\text{merged}}^{(l)}$ is taken from $B_{\text{spec}}^{(l)}$ if the binary mask activates the 
language-specific route for that rank; otherwise, it is taken from $B_{\text{shared}}$.
This yields an effective rank-$r$ adapter whose shared vs.\ language-specific columns are 
controlled by the binary mask $\mathbf{s}^{(l)}$. Thus, Hard routing makes a discrete choice for each rank: 
each column is either fully shared or fully language-specific. During gradient backpropagation, we adopt the straight-through 
estimator (STE) \cite{STE} to approximate gradients through the non-differentiable thresholding 
operation in Eq.~(\ref{eq:zipper_hard_mask}).

\subsection{Zipper-LoRA-Soft (Dynamic Rank-wise Column Mixing)}
\label{subsec:sippersoftdy}

Zipper-LoRA-Soft can be viewed as a continuous relaxation of Zipper-LoRA-Hard. It replaces binary 
selection with \emph{continuous} rank-wise mixing between shared and language-specific columns, 
yielding smoother adaptation.
As shown in Fig.~\ref{fig:zipper_lora}(c), while maintaining the same rank-$r$ bank 
structure and the same router output as Zipper-LoRA-Hard, we directly use the continuous 
mixing vector $\mathbf{p}^{(l)} \in [0, 1]^r$ generated by the language-aware router instead of thresholding it.
We interpret $\mathbf{p}^{(l)}$ as the rank-wise proportion assigned to the language-specific columns, 
and construct the merged up-projection via a weighted summation:
\begin{equation}
  B_{\text{merged}}^{(l)}
  \;=\;
  B_{\text{shared}}\,\mathrm{diag}\!\big(\mathbf{1}-\mathbf{p}^{(l)}\big)
  \;+\;
  B_{\text{spec}}^{(l)}\,\mathrm{diag}\!\big(\mathbf{p}^{(l)}\big)
  \label{eq:zipper_soft}
\end{equation}
where $\mathrm{diag}(\mathbf{p})$ denotes the diagonal matrix with $\mathbf{p}$ on its diagonal, 
i.e., it performs rank-wise (column-wise) scaling.
Equivalently, for each rank index $i \in \{1,\dots,r\}$,
\begin{equation}
  \mathbf{b}_{\text{merged},i}^{(l)}
  \;=\;
  \big(1-p_i^{(l)}\big)\,\mathbf{b}_{\text{shared},i}
  \;+\;
  p_i^{(l)}\,\mathbf{b}_{\text{spec},i}^{(l)}
  \label{eq:zipper_soft_col}
\end{equation}

Compared to Zipper-LoRA-Hard, Zipper-LoRA-Soft provides a continuous trade-off between 
shared and language-specific capacity without requiring discrete rank selection.
Here ``dynamic'' refers to rank-wise adaptive weighting: different rank components 
can be assigned different mixing weights for each language. In summary, both Hard and Soft routing 
use the same language-conditioned router output. Hard routing discretizes this signal to make an 
all-or-nothing column selection, whereas Soft routing keeps it continuous to interpolate between 
shared and language-specific columns.

\subsection{LID-Aware Contextual Routing}
\label{sec:lid_router}

\textbf{Whisper LID Embedding}:
In Fig.~\ref{fig:zipper_lora}, both Zipper-LoRA-Hard and Zipper-LoRA-Soft require a 
language-aware router for LoRA rank selection. We obtain the language identity (LID) 
embedding from the Whisper decoder \cite{whisper}. Specifically, Whisper produces a fixed-dimensional LID 
representation $\mathbf{e}^{(l)} \in \mathbb{R}^{d_{\text{lid}}}$ for target language $l$, 
which serves as a compact language-conditioned signal for routing.

\textbf{Robustness to Imperfect LID}:
The language-aware router uses Whisper LID embeddings as conditioning information, 
but it does not require a perfectly discrete or error-free LID decision. 
The LID embedding continuously modulates the shared and language-specific LoRA rank components 
rather than directly determining the final ASR prediction. As a result, Zipper-LoRA-Soft can 
mitigate minor LID uncertainty through smooth rank-wise composition, since shared and specific 
components are continuously mixed. However, severe LID errors may still guide the router toward 
inappropriate language-specific components and weaken the effect of decoupling. This limitation 
can be alleviated by retaining shared rank components, using soft routing when LID reliability is 
uncertain, and, in future work, incorporating noisy or top-$k$ LID training to improve robustness 
under uncertain language identification.

\textbf{Router Architecture}:
As illustrated by the router block in Fig.~\ref{fig:zipper_lora}, we adopt a lightweight router 
that maps the Whisper LID embedding to rank-wise mixing weights.
Specifically, we first normalize the LID embedding and then project it to the LoRA rank dimension:
\begin{equation}
  \tilde{\mathbf{e}}^{(l)}
  = \mathrm{LayerNorm}\!\left(\mathbf{e}^{(l)}\right),
  \label{eq:lid_ln}
\end{equation}
\begin{equation}
  \hat{\mathbf{e}}^{(l)}
  = W_r\,\tilde{\mathbf{e}}^{(l)} + \mathbf{b}_r ,
  \quad
  \hat{\mathbf{e}}^{(l)} \in \mathbb{R}^{r}
  \label{eq:lid_linear}
\end{equation}
where $W_r \in \mathbb{R}^{r \times d_{\text{lid}}}$ and $\mathbf{b}_r \in \mathbb{R}^{r}$ are trainable parameters.
Then, we apply an element-wise sigmoid to obtain rank-wise mixing weights:
\begin{equation}
  \mathbf{p}^{(l)} = \sigma\!\left(\hat{\mathbf{e}}^{(l)}\right),
  \quad
  \mathbf{p}^{(l)} \in [0,1]^r
  \label{eq:lid_gate}
\end{equation}
The resulting $\mathbf{p}^{(l)}$ is the router output shown in Fig.~\ref{fig:zipper_lora}(b) and (c), 
and it directly controls rank-wise composition between the shared and language-specific banks.
In particular, Zipper-LoRA-Soft uses $\mathbf{p}^{(l)}$ as continuous mixing weights in Eq.~\eqref{eq:zipper_soft}, while Zipper-LoRA-Hard derives a binary mask $\mathbf{s}^{(l)} \in \{0,1\}^r$ by thresholding $\mathbf{p}^{(l)}$ (Eq.~\eqref{eq:zipper_hard_mask}) and performs rank-wise selection via Eq.~\eqref{eq:zipper_hard_zip}.

\subsection{Training Strategy of Speech-LLM with Zipper-LoRA}
\label{subsec:trainstra}

This section describes how the model is trained for multilingual ASR adaptation and 
how the proposed initialization and efficiency strategies are applied. Stage~1 serves as 
a foundation baseline to obtain a well-calibrated Speech-LLM backbone. 
All subsequent experiments, including different LoRA variants, are conducted by inserting 
LoRA modules into the speech encoder and training them in a parameter-efficient supervised 
fine-tuning (SFT) regime.

\textbf{Stage 1: Foundation Baseline}:
In Stage 1, we train an alignment baseline to improve the coupling between the speech encoder 
and the LLM. Specifically, as the  pipeline used in \cite{universal_lora}, 
we update the speech encoder parameters and the modality projector, 
and train LoRA parameters injected into the LLM, while keeping the remaining LLM backbone frozen. 
This stage provides a strong starting point with improved cross-modal alignment, and is used as the 
backbone for Stage~2. Notably, Stage~1 does \emph{not} involve encoder-side LoRA; encoder LoRA is only 
introduced in Stage~2 for all PEFT variants.

\textbf{Stage~2: Parameter-Efficient Multilingual ASR SFT}: 
In Stage~2, we freeze the speech encoder weights obtained from Stage~1 and insert LoRA modules into the 
encoder. We then perform multilingual SFT by updating only the encoder-side LoRA 
parameters (with different variants), the projector, and the LLM-side LoRA parameters. Under this 
setting, different PEFT frameworks differ only in how the encoder-side low-rank update is structured and 
shared across languages, while the rest of the training pipeline remains identical.

\textbf{Segment-wise (Chunked) Encoder Processing for Training Efficiency}:
Training on long-form utterances with full-context encoding can be memory and compute-intensive. To 
improve training throughput, we adopt a segment-wise (chunked) encoder setting. Given an utterance, we 
split the input audio into fixed-length segments and encode each segment independently using the same 
encoder. Segment representations are concatenated in temporal order and fed into the projector. This 
strategy reduces peak GPU memory usage while keeping the Speech-LLM architecture unchanged. In 
experiments, we report results under both full-context and segment-wise encoder regimes to verify 
robustness to encoder input processing.

\textbf{Initial-B Warm-start Initialization}:
To stabilize optimization of rank-wise routing under long-tailed multilingual supervision, we propose an 
\textit{Initial-B warm-start} strategy. In standard LoRA, the up-projection matrix $B$ is typically 
initialized to zeros, which can delay effective adaptation and make joint learning of routing and 
multilingual specialization less stable. Instead of zero initialization, we warm-start the $B$ 
parameters using a converged Zipper-LoRA model.

Specifically, we first train a Zipper-LoRA variant with rank-wise mixing under the Stage~2 SFT 
setting to obtain a converged solution. We then reuse its learned up-projection parameters to initialize 
subsequent runs: both the shared bank $B_{\text{shared}}$ and the language-specific bank 
$B_{\text{spec}}^{(l)}$ are initialized from the pretrained Zipper-LoRA parameters (and the router 
can optionally be warm-started as well). This warm-start provides a well-shaped low-rank subspace at the 
beginning of training, enabling the model to immediately leverage meaningful rank components and 
improving convergence and training stability for the target setting.

\section{EXPERIMENTAL CONFIGURATIONS}
\label{sec:exp}

\subsection{Datasets}
\label{subsec:dataset}

\begin{table}[t]
  \centering
  \caption{Dataset statistics (duration in hours).}
  \label{tab:dataset_stats}
  \renewcommand{\arraystretch}{1.15}%
  \resizebox{\linewidth}{!}{
    \begin{tabular}{llccc}
      \toprule
      \textbf{Setting} & \textbf{Dataset} & \textbf{Language} & \textbf{Train(h)}  & \textbf{Test}(h)\\
      \midrule
      \multirow[c]{4}{*}{\begin{tabular}{@{}c@{}}Baseline\\(source)\end{tabular}}
      & \multirow{3}{*}{MSR86k\cite{msr86k}} & English (en) & 3000  & 14.72 \\
      &                        & French (fr)  & 3000 & 12.60  \\
      &                        & Thai (th)    & 3000  & 8.37 \\
      \cmidrule(lr){2-5}
      & WenetSpeech (WS)\cite{wenetspeech}            & Chinese (zh) & 3000  & 15.18 $\mid$ 23.06 \\
      \midrule
      \multirow[c]{12}{*}{\begin{tabular}{@{}c@{}}SFT\\(target)\end{tabular}}
      & \multirow{4}{*}{MLS\cite{mls}}   & German (de)  & 1000 & 14.29  \\
      &                        & Spanish (es) & 1000 & 10.00 \\
      &                        & French (fr)  & 1000 & 10.07 \\
      &                               & Italian (it)    & 500 & 8.10  \\
      \cmidrule(lr){2-5}
      & \multirow{2}{*}{MSR86k\cite{msr86k}}   & Russian (ru) & 1000 & 8.16  \\
      &                        & Vietnamese (vi) & 1000 & 7.25 \\
      \cmidrule(lr){2-5}
      & LibriSpeech (LS)\cite{librispeech}                    & English (en)   & 960  & 5.40 $\mid$ 5.34  \\
      \cmidrule(lr){2-5}
      & \multirow{5}{*}{Common Voice (CV)\cite{common_voice}} & Thai (th)     & 865 & 4.00  \\
      &                               & Arabic (ar)   & 1 & 4.00 \\
      &                               & Japanese (ja) & 1 & 4.00 \\
      &                               & Korean (ko)   & 1 & 0.98 \\
      &                               & Portuguese (pt) & 1 & 1.00 \\
      \bottomrule
    \end{tabular}
  }
  \vspace{1mm}
  \parbox{0.98\linewidth}{\footnotesize
  \textit{Note:} “$\mid$” separates two test subsets.
  For WenetSpeech, 15.18 $\mid$ 23.06 correspond to Meeting and Net.
  For LibriSpeech, 5.40 $\mid$ 5.34 correspond to Clean and Other.}
\end{table}

Table~\ref{tab:dataset_stats} summarizes the data used in our two-stage training. 
The foundation baseline stage uses high-resource corpora to build a multilingual ASR initialization, 
while the SFT stage (stage 2) adapts the model to a 12-language setting with 
substantial training resource imbalance.

As shown in Table~\ref{tab:dataset_stats} (Baseline block), we construct the foundation baseline 
training set with four high-resource languages. Specifically, MSR86k provides English, French, and Thai, 
while WenetSpeech provides Chinese; for each language, we randomly sample 3000 hours 
to control the training budget. In the SFT block, we perform multilingual ASR SFT adaptation  
on 12 target languages with a long-tailed distribution. The higher-resource portion consists of MLS 
German, Spanish, and French (1000 hours each) and Italian (500 hours), MSR86k Russian and Vietnamese 
(1000 hours each), LibriSpeech English (960 hours), and Common Voice Thai (865 hours), with all hours 
following the sampling setup listed in the table. The lowest-resource portion consists of Common Voice 
Arabic, Japanese, Korean, and Portuguese with 1 hour per language. Some languages overlap across stages but come 
from different corpora (French, Thai, and English), introducing domain mismatch that can 
increase cross-lingual interference under long-tailed SFT adaptation.

For evaluation, we follow the official splits of each dataset whenever available. MSR86k provides an 
official dev split (and no official test split), so we evaluate on its dev set. For MLS, LibriSpeech, and 
WenetSpeech, we evaluate on the official test splits. For Common Voice, we use randomly sampled 
evaluation subsets to control the evaluation scale: Arabic and Thai use 4-hour subsets randomly sampled 
from the official test split, Korean uses the official test split (0.98 hours), and Portuguese uses a
1-hour subset randomly sampled from the official training split as a held-out evaluation set due to the
limited data size and the absence of an official test split. All random sampling procedures are fixed and 
shared across all methods to ensure fair comparisons, and we have released the exact utterance lists 
for reproducibility.

\subsection{Configurations}
\label{subsec:config}

We use Whisper Large-v3~\cite{whisper} as the speech encoder and Qwen3-1.7B~\cite{yang2025qwen3} 
as the decoder-only language model. The modality projector adopts a gated projection module 
that performs temporal downsampling by a factor of $4$. Specifically, it applies two 1D convolution 
layers (kernel size $4$, stride $4$) to produce a gate branch and an up-projection branch, 
fuses them via a SiLU-gated interaction \cite{silu}, and then applies a linear transformation 
followed by a residual connection with layer normalization.  A final linear projection outputs 
the speech-conditioned embedding sequence. Combined with the initial $2\times$ downsampling 
in the speech encoder, this results in an overall $8\times$ temporal reduction, yielding a 
highly compressed representation at $12.5$ Hz that is fed into the language model.

For parameter-efficient adaptation, we insert low-rank adaptation modules into all 
linear layers on both the encoder and language-model sides, including the attention 
projections and feed-forward network layers. We set the rank to $32$ and the scaling 
factor to $64$ for all methods. For Zipper-LoRA-Hard, the rank-selection threshold 
$\tau$ is set to $0.5$. For FlyLoRA~\cite{flylora2025}, we activate $8$ rank 
components. For the chunked encoder regime, we randomly sample chunk durations 
from $\{1,2,4,8\}$ seconds during training and use a fixed chunk duration of $8$ seconds 
during decoding; no overlap is used. The non-chunked regime encodes each utterance 
with full context. All experiments are trained on $4$ NVIDIA A800 GPUs (80\,GB) with 
DeepSpeed ZeRO stage $2$ \cite{deepspeed}, using a learning rate of $2\times10^{-5}$, 
a warmup ratio of $10\%$, and a cosine learning-rate schedule. To improve memory 
efficiency, we pack training samples by constraining the total sequence length to at 
most 2048 tokens, where the packed length includes the downsampled speech tokens, 
the text tokens, and reserved prompt tokens; packing is performed within the same 
language. After packing, we use a batch size of $16$ for chunked training and a 
batch size of $10$ for non-chunked training.

For evaluation, we compute character 
error rate (CER) for Chinese, Japanese, Korean, and Thai, and word error rate (WER) 
for all other languages, with Whisper text normalization. All experiments in this work 
follow these configurations unless otherwise specified.

\section{Results and Discussions}
\label{subsec:rstcmp}

\subsection{Source Domain Results}
\label{subsubsec:srcrst}

\begin{table}[t]
  \centering
  \caption{Foundation baseline WER/CER (\%) comparison under chunked vs. non-chunked settings.}
  \label{tab:baseline_chunk_vs_nonchunk}
  \footnotesize
    \renewcommand{\arraystretch}{1.15}%
    \begin{tabular}{lccccc}
     \toprule
      & \multicolumn{3}{c}{\textbf{MSR86k}} & \multicolumn{2}{c}{\textbf{WenetSpeech}} \\
      \cmidrule(lr){2-4}\cmidrule(lr){5-6}
      & en & fr & th & Meeting & Net \\
      \midrule
      Chunked      & 4.65 & \textbf{9.24} & 4.20 & \textbf{16.55} & \textbf{9.60} \\
      Non-chunked  & \textbf{4.59} & 9.29 & \textbf{4.05} & 17.16 & 10.22 \\
      \bottomrule
    \end{tabular}
\end{table}

\begin{table*}[t]
  \centering
  \caption{Component configurations of the compared LoRA-based adaptation methods.}
  \label{tab:ablation_config}
  \footnotesize
  \renewcommand{\arraystretch}{1.15}%
  \begin{tabular}{@{}lcccccc@{}}
    \toprule
    \multirow{2}{*}{\textbf{Methods}}
    & \multirow{2}{*}{\textbf{Shared}}
    & \multirow{2}{*}{\textbf{Specific}}
    & \multirow{2}{*}{\textbf{LID Router}}
    & \textbf{Dynamic Rank}
    & \multirow{2}{*}{\textbf{Soft Mixing}}
    & \multirow{2}{*}{\textbf{Initial-B}} \\
    & & & & \textbf{Routing} & & \\
    \midrule
    Vanilla-LoRA                 & $\checkmark$ & --           & --           & --           & --           & -- \\
    FlyLoRA                      & $\checkmark$ & --           & --           & $\checkmark$ & --           & -- \\
    Independent-LoRA             & --           & $\checkmark$ & --           & --           & --           & -- \\
    \midrule
    Zipper-LoRA-Static           & $\checkmark$ & $\checkmark$ & --           & --           & --           & -- \\
    Zipper-LoRA-Hard             & $\checkmark$ & $\checkmark$ & $\checkmark$ & $\checkmark$ & --           & -- \\
    Zipper-LoRA-Soft             & $\checkmark$ & $\checkmark$ & $\checkmark$ & $\checkmark$ & $\checkmark$ & -- \\
    Zipper-LoRA-Soft + Initial-B & $\checkmark$ & $\checkmark$ & $\checkmark$ & $\checkmark$ & $\checkmark$ & $\checkmark$ \\
    \bottomrule
  \end{tabular}

  \vspace{0.3em}
  \parbox{0.98\linewidth}{\footnotesize\textit{Note:}
  ``Shared'' and ``Specific'' indicate shared and language-specific adaptation capacity, respectively.
  FlyLoRA performs input-dependent implicit top-$k$ dynamic rank routing, whereas the Zipper-LoRA router is explicitly conditioned on language-ID embeddings.}
\end{table*}

\begin{table}[t]
  \centering
  \caption{Source-domain performance after SFT on MSR86k (en/fr/th) and WenetSpeech (Meeting/Net) under chunked and non-chunked encoder settings.}
  \label{tab:source_domain}
  \resizebox{\linewidth}{!}{
    \begin{tabular}{lccccc}
      \toprule
      \multirow{2}{*}{\textbf{Methods}} & \multicolumn{3}{c}{\textbf{MSR86k}} & \multicolumn{2}{c}{\textbf{WenetSpeech}} \\
      \cmidrule(lr){2-4}\cmidrule(lr){5-6}
      & en & fr & th & Meeting & Net \\
      \midrule
      \multicolumn{6}{l}{\textbf{Chunked}} \\
      \midrule
      Vanilla-LoRA                        & 4.88 & 10.74 & 13.58 & 18.51 & 12.12 \\
      FlyLoRA                    & 4.87 & \textbf{10.65} & 7.99  & 17.57 & 11.86 \\
      Independent-LoRA            & 4.86 & 11.07 & 7.72  & \textbf{17.24} & 11.63 \\
      Zipper-LoRA-Static          & 4.91 & 11.35 & 7.85  & 18.12 & 12.11 \\
      Zipper-LoRA-Hard            & \textbf{4.79} & 11.06 & \textbf{7.59}  & 17.49 & \textbf{11.61} \\
      Zipper-LoRA-Soft            & 4.88 & 11.26 & 7.79  & 17.90 & 11.93 \\
      \quad \quad \quad + Initial-B  & 5.06 & 11.53 & 8.05  & 17.96 & 12.02 \\
      \midrule
      \multicolumn{6}{l}{\textbf{Non-Chunked}} \\
      \midrule
      Vanilla-LoRA                        & \textbf{4.40} & \textbf{10.10} & 12.04 & 18.40 & 12.38 \\
      FlyLoRA                    & 4.43 & \textbf{10.10} & \textbf{7.10}  & 18.25 & 12.31 \\
      Independent-LoRA            & 4.59 & 10.86    & 7.50    & \textbf{17.58} & \textbf{11.76} \\
      Zipper-LoRA-Static          & 4.64 & 11.07 & 7.70  & 17.68 & 11.97 \\
      Zipper-LoRA-Hard            & 4.53 & 10.78 & 7.40  & 18.06 & 11.79 \\
      Zipper-LoRA-Soft            & 4.59 & 11.02 & 7.52  & 17.99 & 11.90 \\
      \quad \quad \quad + Initial-B  & 4.69 & 11.24 & 7.69  & 17.83 & 11.94 \\
      \bottomrule
    \end{tabular}
  }

  \vspace{1mm}
  \parbox{0.98\linewidth}{\footnotesize\textit{Note:} ``+ Initial-B'' denotes warm-start for Zipper-LoRA-Soft.}
\end{table}

\begin{figure}[t]
  \centering
  \includegraphics[width=0.8\linewidth]{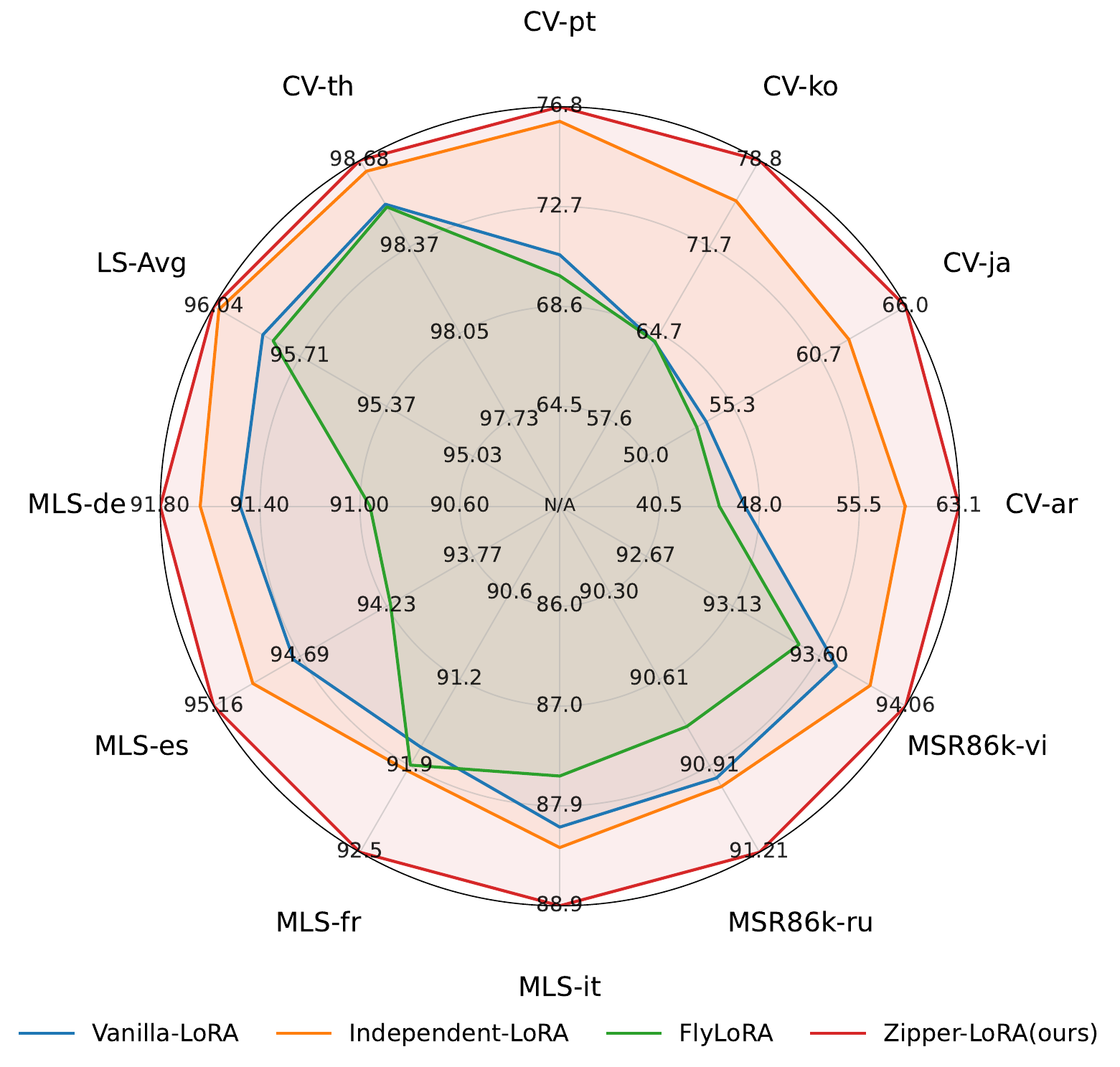}
  \caption{Performance comparison of Zipper-LoRA-Soft + Initial-B and other LoRA-based methods on the 12 target languages in the SFT stage under the non-chunked setting, using $(1-\mathrm{WER/CER})\%$ as the metric. Detailed numerical results are provided in Table~\ref{tab:target_radar_raw}.}
  \label{fig:radar}
\end{figure}

Table~\ref{tab:baseline_chunk_vs_nonchunk} reports pretrained-stage foundation baseline ASR performance under chunked and non-chunked encoder settings. The two settings achieve comparable WER/CER on both MSR86k and WenetSpeech. These results suggest that the chunked encoder does not introduce an evident degradation in recognition performance at the pretraining stage. Under our training setup, chunking also leads to a substantial improvement in training efficiency.

Table~\ref{tab:source_domain} presents the source-domain ASR performance after SFT adaptation under both chunked and non-chunked encoder configurations. A comparison between the foundation baseline (Table~\ref{tab:baseline_chunk_vs_nonchunk}) and the post-SFT results reveals a general performance degradation across all evaluated languages. This decline is primarily caused by domain mismatch and cross-lingual interference during the second-stage adaptation. Although languages such as English, French, and Thai are present in both stages, the SFT stage uses different corpora (e.g., LibriSpeech and Common Voice) compared to the foundation stage (MSR86k). This introduces significant distributional shifts; for instance, in the Vanilla-LoRA (Chunked) setting, the error rate for Thai increases sharply from 4.20\% to 13.58\%. Such results suggest that the long-tailed distribution of the 12-language SFT task leads to severe interference, resulting in ``catastrophic forgetting'' of the high-resource knowledge learned during the foundation stage.

The degree of forgetting varies significantly across different PEFT strategies. 
Vanilla-LoRA shows the most obvious performance drop, particularly on the Thai and WenetSpeech tasks, 
indicating that a shared low-rank space is not enough to protect source-domain knowledge from the noise of 
multi-target adaptation. Independent-LoRA and FlyLoRA reduce this forgetting through greater parameter
decoupling. In particular, Independent-LoRA shows good stability on the WenetSpeech Meeting and Net sets,
although neither method maintains optimal performance across all source-domain tasks.

Our proposed Zipper-LoRA framework, especially the Zipper-LoRA-Hard variant, 
shows better performance in managing the trade-off between target-domain adaptation and source-domain preservation. 
In the chunked encoder setting, Zipper-LoRA-Hard achieves the lowest error rates for English (4.79\%), Thai (7.59\%), 
and WenetSpeech Net (11.61\%) among all compared SFT methods. This performance gain highlights 
the effectiveness of the Zipper-LoRA routing mechanism in isolating interference from the long-tailed target 
distribution. Furthermore, the way Zipper-LoRA and chunked encoding work together is clear 
in long-form scenarios like WenetSpeech, where it consistently performs better than non-chunked 
alternatives. These results confirm that Zipper-LoRA-Hard provides a strong protection, 
effectively ``zipping" the foundation and adaptation layers to reduce the negative impact of domain mismatch.
\subsection{Target Domain Results}
\label{subsec:tgtrst}

\begin{table*}[t]
  \centering
  \caption{WER/CER (\%) on the 12 target languages in the SFT stage under the non-chunked encoder setting.}
  \label{tab:target_radar_raw}
  \footnotesize
  \renewcommand{\arraystretch}{1.15}%
  \setlength{\tabcolsep}{2.7pt}%
  \sbox{\sigtablebox}{%
  \begin{tabular}{@{}l ccccc c cccc cc@{}}
    \toprule
    \multirow{2}{*}{\textbf{Methods}}
    & \multicolumn{5}{c}{\textbf{CommonVoice}}
    & \multicolumn{1}{c}{\textbf{LibriSpeech}}
    & \multicolumn{4}{c}{\textbf{MLS}}
    & \multicolumn{2}{c}{\textbf{MSR86k}} \\
    \cmidrule(lr){2-6}\cmidrule(lr){7-7}\cmidrule(lr){8-11}\cmidrule(lr){12-13}
    & ar & ja & ko & pt & th
    & Avg
    & de & es & fr & it
    & ru & vi \\
    \midrule
    Vanilla-LoRA     & 53.08 & 46.28 & 36.03 & 29.30 & 1.48 & 4.15 & 8.52 & 5.27 & 8.26 & 11.86 & 9.05 & 6.31 \\
    FlyLoRA          & 55.01 & 46.86 & 35.99 & 30.16 & 1.49 & 4.19 & 9.04 & 5.79 & 8.13 & 12.36 & 9.23 & 6.51 \\
    Independent-LoRA & 41.01 & 37.53 & 24.53 & 23.82 & 1.36 & 3.98 & 8.36 & 5.05 & 8.09 & 11.66 & 9.02 & 6.13 \\
    Zipper-LoRA-Soft + Initial-B & \textbf{36.94} & \textbf{34.04} & \textbf{21.35} & \textbf{23.22} & \textbf{1.32} & \textbf{3.96} & \textbf{8.20} & \textbf{4.84} & \textbf{7.50} & \textbf{11.09} & \textbf{8.79} & \textbf{5.94} \\
    \bottomrule
  \end{tabular}%
  }
  \usebox{\sigtablebox}

  \vspace{0.3em}
  \parbox{\wd\sigtablebox}{\footnotesize\textit{Note:} All results are the detailed WER/CER values corresponding to the visualization in Fig.~\ref{fig:radar}.}
\end{table*}

Table~\ref{tab:target_radar_raw} reports the detailed WER/CER results for multilingual ASR adaptation
on the 12 target languages in the supervised fine-tuning stage under the non-chunked encoder setting. 
Zipper-LoRA-Soft + Initial-B achieves the lowest error rate on all 12 target languages.
The gains are especially clear on low-resource Common Voice languages, where it reduces WER/CER 
from 53.08\% to 36.94\% on Arabic, from 46.28\% to 34.04\% on Japanese, and from 36.03\% to 21.35\% 
on Korean compared with Vanilla-LoRA. It also consistently improves over Independent-LoRA and FlyLoRA 
across both low-resource and high-resource target languages. These results suggest that rank-level 
composition with controlled sharing effectively mitigates inter-lingual interference while preserving 
cross-lingual transferable acoustic representations. Fig.~\ref{fig:radar} further provides a visual 
supplement to Table~\ref{tab:target_radar_raw} by converting the WER/CER values into 
$(1-\mathrm{WER/CER})\%$ scores, where the larger enclosed area of Zipper-LoRA-Soft + Initial-B
indicates more balanced performance across languages.

\subsubsection{Results on high-resource target languages}
\label{subsubsec:hightgt}

\begin{table*}[t]
  \centering
  \caption{WER/CER (\%) on high-resource target languages under chunked and non-chunked encoder settings.}
  \label{tab:wer_high}
  \footnotesize
  \renewcommand{\arraystretch}{1.15}%
  \sbox{\sigtablebox}{%
  \begin{tabular}{@{}l c ccc cccc cc@{}}
    \toprule
    \multirow{2}{*}{\textbf{Methods}}
    & \multicolumn{1}{c}{\textbf{CommonVoice}}
    & \multicolumn{3}{c}{\textbf{LibriSpeech}}
    & \multicolumn{4}{c}{\textbf{MLS}}
    & \multicolumn{2}{c}{\textbf{MSR86k}} \\
    \cmidrule(lr){2-2}\cmidrule(lr){3-5}\cmidrule(lr){6-9}\cmidrule(lr){10-11}
    & th
    & Clean & Other & Avg
    & de & es & fr & it
    & ru & vi \\
    \midrule
    \multicolumn{11}{@{}l}{\textbf{Chunked}} \\
    \midrule
    Vanilla-LoRA      & 1.47 & 2.77 & 5.88 & 4.32 & 9.57 & 6.31 & 8.31 & 12.57 & 9.59 & 6.64 \\
    FlyLoRA           & 1.57 & 2.80 & 5.90 & 4.35 & 9.71 & 6.48 & 8.28 & 12.92 & 9.70 & 6.75 \\
    Independent-LoRA  & 1.30 & 2.60 & 5.63 & 4.11 & 9.08 & 5.67 & 8.02 & 11.65 & 9.40 & 6.33 \\
    Zipper-LoRA-Static & 1.35 & 2.73 & 5.58 & 4.15 & 9.20 & 6.10 & 8.15 & 11.87 & 9.31 & 6.41 \\
    Zipper-LoRA-Hard   & 1.35 & 2.62 & 5.53 & 4.07 & 9.15 & 5.69 & 8.03 & 11.71 & 9.44 & 6.36 \\
    Zipper-LoRA-Soft   & 1.31 & 2.61 & 5.54 & 4.07 & 9.15 & 5.62 & 8.03 & 11.75 & 9.37 & 6.32 \\
    \quad + Initial-B
    & \textbf{1.26}$^{\dagger}$
    & \textbf{2.56}$^{\dagger}$
    & \textbf{5.45}$^{\dagger}$
    & \textbf{4.00}$^{\dagger}$
    & \textbf{8.93}$^{\dagger\ddagger}$
    & \textbf{5.31}$^{\dagger\ddagger}$
    & \textbf{7.90}$^{\dagger\ddagger}$
    & \textbf{11.10}$^{\dagger\ddagger}$
    & \textbf{9.21}$^{\dagger\ddagger}$
    & \textbf{6.15}$^{\dagger\ddagger}$ \\
    \midrule
    \multicolumn{11}{@{}l}{\textbf{Non-Chunked}} \\
    \midrule
    Vanilla-LoRA      & 1.48 & 2.66 & 5.65 & 4.15 & 8.52 & 5.27 & 8.26 & 11.86 & 9.05 & 6.31 \\
    FlyLoRA          & 1.49 & 2.70 & 5.68 & 4.19 & 9.04 & 5.79 & 8.13 & 12.36 & 9.23 & 6.51 \\
    Independent-LoRA & 1.36 & 2.53 & 5.44 & 3.98 & 8.36 & 5.05 & 8.09 & 11.66 & 9.02 & 6.13 \\
    Zipper-LoRA-Static & \textbf{1.30} & \textbf{2.52} & 5.45 & 3.98 & 8.53 & 5.00 & 7.96 & 11.95 & 8.97 & 6.18 \\
    Zipper-LoRA-Hard   & 1.35 & 2.56 & 5.49 & 4.02 & 8.39 & 4.86 & 7.82 & 11.36 & 8.94 & 6.19 \\
    Zipper-LoRA-Soft   & 1.36 & 2.55 & 5.47 & 4.01 & 8.38 & 4.86 & 7.84 & 11.41 & 8.87 & 6.11 \\
    \quad + Initial-B & 1.32$^{\dagger}$ & 2.53$^{\dagger}$ & \textbf{5.38}$^{\dagger\ddagger}$ & \textbf{3.96}$^{\dagger\ddagger}$ & \textbf{8.20}$^{\dagger\ddagger}$ & \textbf{4.84}$^{\dagger\ddagger}$ & \textbf{7.50}$^{\dagger\ddagger}$ & \textbf{11.09}$^{\dagger\ddagger}$ & \textbf{8.79}$^{\dagger\ddagger}$ & \textbf{5.94}$^{\dagger\ddagger}$ \\
    \bottomrule
  \end{tabular}%
  }
  \usebox{\sigtablebox}

  \vspace{0.3em}
  \parbox{\wd\sigtablebox}{\footnotesize\textit{Note:}
  The component settings are summarized in Table~\ref{tab:ablation_config}.
  The superscript markers $^{\dagger}$ and $^{\ddagger}$ denote that Zipper-LoRA-Soft + Initial-B is significantly better than Vanilla-LoRA and Independent-LoRA, respectively. Significance is tested using paired bootstrap resampling \cite{bisani2004bootstrap,koehn2004statistical} with 1,000 bootstrap samples ($p<0.05$).}
\end{table*}

\begin{table}[t]
  \centering
  \caption{WER/CER (\%) on low-resource Common Voice languages under chunked and non-chunked settings.}
  \label{tab:wer_low_cv}
  \footnotesize
  \begin{tabular}{@{}lcccc@{}}
    \toprule
    \textbf{Methods} & ar & ja & ko & pt \\
    \midrule
    \multicolumn{5}{@{}l}{\textbf{Chunked}} \\
    \midrule
    Vanilla-LoRA        & 62.32 & 48.93 & 38.37 & 33.49 \\
    FlyLoRA             & 65.20 & 49.75 & 39.02 & 33.58 \\
    Independent-LoRA    & 45.24 & 37.84 & 25.98 & 25.15 \\
    Zipper-LoRA-Static  & 42.36 & 37.51 & 25.65 & \textbf{23.26} \\
    Zipper-LoRA-Hard    & 43.07 & 37.10 & 25.80 & 23.86 \\
    Zipper-LoRA-Soft    & 41.72 & 35.77 & 24.37 & 23.89 \\
    \quad + Initial-B   & \textbf{38.64}$^{\dagger\ddagger}$ & \textbf{35.21}$^{\dagger\ddagger}$ & \textbf{21.87}$^{\dagger\ddagger}$ & 24.29$^{\dagger}$ \\
    \midrule
    \multicolumn{5}{@{}l}{\textbf{Non-Chunked}} \\
    \midrule
    Vanilla-LoRA        & 53.08 & 46.28 & 36.03 & 29.30 \\
    FlyLoRA             & 55.01 & 46.86 & 35.99 & 30.16 \\
    Independent-LoRA    & 41.01 & 37.53 & 24.53 & 23.82 \\
    Zipper-LoRA-Static  & 40.32 & 36.61 & 23.39 & 24.61 \\
    Zipper-LoRA-Hard    & 40.89 & 36.55 & 24.61 & 23.96 \\
    Zipper-LoRA-Soft    & 40.07 & 35.66 & 22.55 & 23.47 \\
    \quad + Initial-B   & \textbf{36.94}$^{\dagger\ddagger}$ & \textbf{34.04}$^{\dagger\ddagger}$ & \textbf{21.35}$^{\dagger\ddagger}$ & \textbf{23.22}$^{\dagger}$ \\
    \bottomrule
  \end{tabular}

  \vspace{0.3em}
  \parbox{\linewidth}{\footnotesize\textit{Note:}
  The component settings are summarized in Table~\ref{tab:ablation_config}.
  The superscript markers $^{\dagger}$ and $^{\ddagger}$ denote statistical significance tests. Specifically, $^{\dagger}$ represents that Zipper-LoRA-Soft + Initial-B is significantly better than Vanilla-LoRA, and $^{\ddagger}$ represents that Zipper-LoRA-Soft + Initial-B is significantly better than Independent-LoRA. Significance is tested using paired bootstrap resampling with 1,000 bootstrap samples ($p<0.05$).}
\end{table}

Table~\ref{tab:ablation_config} explicitly summarizes which components are enabled for each method, while
Table~\ref{tab:wer_high} reports SFT ASR performance on high-resource target languages under both
the chunked and non-chunked encoder settings, 
including Common Voice (Thai), LibriSpeech (Clean, Other and Average), Multilingual LibriSpeech 
(German, Spanish, French and Italian) and MSR86k (Russian and Vietnamese). The two encoder settings
are explicitly presented as the Chunked and Non-Chunked row blocks in Table~\ref{tab:wer_high}.
Across both settings, the Zipper-LoRA family consistently delivers strong performance. Taken together,
the component configurations in Table~\ref{tab:ablation_config} and the performance results in
Table~\ref{tab:wer_high} reveal a clear progressive ablation pattern, with four findings standing out.
1) Zipper-LoRA-Static performs competitively with Independent-LoRA,
indicating that the rank-wise shared/specific decomposition can already balance cross-lingual sharing
and language-specific adaptation. 2) Zipper-LoRA-Hard further improves or matches the static variant on
several languages, showing the contribution of LID-aware binary rank selection. Zipper-LoRA-Soft
achieves comparable or better performance than Zipper-LoRA-Hard on several languages, suggesting that 
continuous rank-wise mixing provides a smoother trade-off between shared and language-specific subspaces. 
In particular, Zipper-LoRA-Soft closely matches or surpasses Independent-LoRA on several high-resource 
languages, such as Spanish, French, Italian, Russian, and Vietnamese under the non-chunked setting, 
indicating that continuous rank-wise mixing can serve as an effective alternative to full language-specific 
decoupling. 3) Moreover, the + Initial-B warm-start further improves Zipper-LoRA-Soft and achieves
statistically significant gains over the main baselines on most high-resource evaluations.
4) Overall, the component configurations in Table~\ref{tab:ablation_config} and the progressive
performance trend in Table~\ref{tab:wer_high} show that the improvements arise from the combination of
rank-wise decomposition, language-aware routing, discrete or continuous rank composition, and warm-start
initialization, rather than from a single factor alone. Fig.~\ref{fig:radar} further illustrates the
balanced overall performance of Zipper-LoRA-Soft + Initial-B across the 12 target languages. By contrast,
Vanilla-LoRA and FlyLoRA underperform compared with the proposed Zipper-LoRA variants, likely due to the
inter-lingual interference introduced by parameter sharing.

\subsubsection{Results on low-resource target languages}
\label{subsubsec:lowtgtrst}

In the extremely low-resource adaptation tasks (only 1 hour of fine-tuning data per language), 
Table~\ref{tab:wer_low_cv} further highlights the ablation effect of the proposed variants. 
Five notable trends emerge from the low-resource results in Table~\ref{tab:wer_low_cv}.
1) All Zipper-LoRA variants achieve lower WER/CER than Vanilla-LoRA on the four lowest-resource
Common Voice languages under both chunked and non-chunked settings, indicating that structured cross-lingual sharing is crucial when supervision
is extremely limited. 2) Zipper-LoRA-Static already improves over Vanilla-LoRA and also outperforms
Independent-LoRA on several languages, showing that the rank-wise shared/specific decomposition itself 
is beneficial: it preserves transferable cross-lingual knowledge while introducing language-specific 
capacity to reduce interference. 3) Compared with the static variant, Zipper-LoRA-Hard and Zipper-LoRA-Soft
achieve further gains in many cases, indicating the additional benefit of language-aware dynamic routing. 
The stronger and more stable performance of Zipper-LoRA-Soft over Zipper-LoRA-Hard suggests that 
continuous weighted composition is more suitable than threshold-based discrete selection for balancing 
shared and language-specific ranks. 4) Finally, Zipper-LoRA-Soft + Initial-B achieves the best overall
low-resource performance and obtains statistically significant improvements over the main baselines in most
comparisons, confirming that the warm-start strategy stabilizes and enhances the proposed architecture. 
5) Overall, these results demonstrate that Zipper-LoRA provides a better trade-off between cross-lingual
knowledge transfer and inter-lingual interference than either fully shared or fully decoupled designs.

\subsubsection{Scalability from extremely-low to low-resource target languages}
\label{subsubsec:visual}

\begin{figure}[t]
  \centering
  \includegraphics[width=0.9\linewidth]{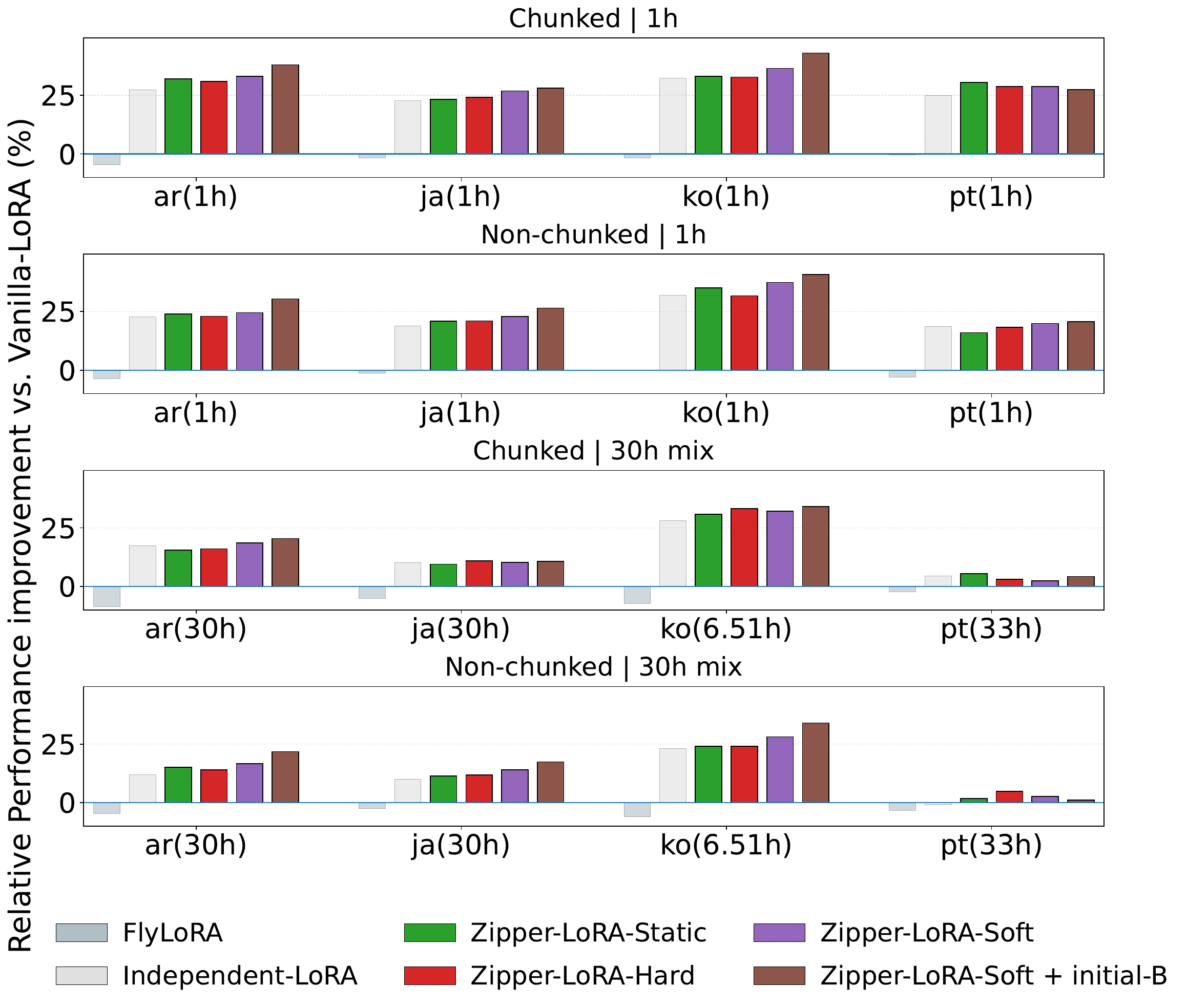}
  \caption{Performance gap visualization from extremely-low to low-resource target languages. 
  Bars show relative performance change vs. Vanilla-LoRA; positive values indicate improvement and negative values indicate 
  degradation. The 1-hour setting uses 1 hour per language as in Table~\ref{tab:wer_low_cv}; 
  for the 30-hour setting, Arabic, Japanese, and Korean are from Common Voice (Korean uses 6.51 hours due to the limited official data size), while Portuguese is from MLS.}
  \label{fig:low_resource_vis}
\end{figure}

Fig.~\ref{fig:low_resource_vis} shows the visualized relative performance change of each method 
compared to Vanilla-LoRA. Positive bars indicate improvements, while negative bars show a drop in performance. 
This visualization examines whether the benefits of the proposed methods hold when the training data 
for low-resource target languages increases from 1 hour (extremely low-resource in Table \ref{tab:wer_low_cv}) 
to 30 hours (low-resource).

Across both chunked and non-chunked regimes, Zipper-LoRA variants consistently show gains over Vanilla-LoRA. 
The most significant improvements are observed in the 1-hour setting, where the lack of data makes structured 
cross-lingual sharing vital for the model to learn effectively. As the data size increases to 30 hours, 
the relative improvement gap narrows slightly. This is expected, as the increased supervision allows 
the model to rely more on language-specific information. However, the Zipper-LoRA-Soft + Initial-B variant
still maintains a clear advantage over all other methods even at the 30-hour condition. 
This suggests that our approach is not just a solution for extreme data scarcity, but remains effective 
as more data becomes available. This is also consistent with the high-resource target-language ASR
results reported in Table~\ref{tab:wer_high}.

In contrast, fully shared adaptations such as FlyLoRA can exhibit smaller gains or
even degradation, suggesting stronger sensitivity to negative knowledge transfer when 
low-resource languages are adapted jointly with higher-resource ones. 
While Independent-LoRA provides strong protection against interference by 
isolating parameters, it cannot capture the cross-lingual benefits that Zipper-LoRA 
achieves through its rank-wise composition. Overall, these results further confirm 
that Zipper-LoRA provides a flexible framework that remains strong across different data scales, 
effectively managing the transition from extremely low to low-resource conditions.

\subsubsection{Robust gains and the effect of encoder context}
\label{subsubsec:robust}

Comparing the Chunked and Non-Chunked row blocks in Table~\ref{tab:wer_high}, together with
Table~\ref{tab:wer_low_cv}, shows that the results are robust to the encoder processing strategy. 
Across all methods, the non-chunked setting consistently yields lower absolute WER 
than the chunked setting, since chunked attention processes each block independently and prevents the 
encoder from attending across chunks, thereby limiting the available contextual information. 
Despite this systematic gap, Zipper-LoRA variants, particularly soft composition and the Initial-B warm-start,
consistently achieve the lowest or near-lowest error rates across both high-resource benchmarks and 
the low-resource Common Voice subset. This consistency supports the claim that rank-wise sharing 
and specialization provide a reliable mechanism for balancing cross-lingual knowledge transfer and 
inter-lingual interference under different encoder context conditions in practical multilingual adaptation.

\subsection{Efficiency and Parameter Analysis}
\label{subsec:efficiency_param}

\begin{table}[t]
  \centering
  \caption{Encoder efficiency under different chunk sizes.}
  \label{tab:encoder_efficiency}
  \small
  \renewcommand{\arraystretch}{1.15}%
  \resizebox{\linewidth}{!}{%
    \begin{tabular}{lcccc}
      \toprule
      \multirow{2}{*}{\textbf{Chunk Size}} & \textbf{Encoder Peak} & \textbf{Encoder Extra Peak} & \textbf{Extra Mem.} & \textbf{Relative} \\
      & \textbf{GPU Mem.} & \textbf{GPU Mem.} & \textbf{Reduction} & \textbf{Speed} \\
      \midrule
      Non-chunked & 6.314 GB & 4.947 GB & -- & 1.00$\times$ \\
      8s chunks    & 4.382 GB & 3.011 GB & 39.14\% & 1.41$\times$ \\
      4s chunks    & 4.010 GB & 2.638 GB & 46.68\% & 1.53$\times$ \\
      2s chunks    & 3.645 GB & 2.273 GB & 54.05\% & 1.56$\times$ \\
      1s chunks    & 3.580 GB & 2.208 GB & 55.36\% & 1.56$\times$ \\
      \bottomrule
    \end{tabular}%
  }
\end{table}

\begin{table}[t]
  \centering
  \caption{Encoder-side parameter comparison of PEFT methods.}
  \label{tab:encoder_param_efficiency}
  \small
  \renewcommand{\arraystretch}{1.15}%
  \resizebox{\linewidth}{!}{%
    \begin{tabular}{lccc}
      \toprule
      \multirow{2}{*}{\textbf{Methods}} & \textbf{Encoder} & \textbf{Encoder PEFT} & \textbf{PEFT Param.} \\
      & \textbf{Params.} & \textbf{Params.} & \textbf{Ratio} \\
      \midrule
      Vanilla-LoRA       & 746.48 M  & 23.60 M  & 3.16\% \\
      FlyLoRA           & 746.48 M  & 23.60 M  & 3.16\% \\
      Independent-LoRA   & 1006.02 M & 283.13 M & 28.14\% \\
      Zipper-LoRA-Static & 811.38 M  & 88.47 M  & 10.90\% \\
      Zipper-LoRA-Hard   & 896.41 M  & 173.53 M & 19.36\% \\
      Zipper-LoRA-Soft   & 896.41 M  & 173.53 M & 19.36\% \\
      \bottomrule
    \end{tabular}%
  }
\end{table}

To further examine the practical efficiency of the chunked encoder, Table~\ref{tab:encoder_efficiency} reports the memory and speed statistics of different encoder-side configurations. The profiling is conducted with a 30-second audio input, batch size 1, bf16 precision, and a forward-backward pass. In the table, ``Encoder Extra Peak GPU Mem.'' denotes the additional peak GPU memory attributed to encoder computation under the same profiling setup, and relative speed is normalized to non-chunked encoding. The results show that the chunked encoder substantially reduces memory usage compared with the non-chunked encoder. The 8s chunk setting, which is used as the default decoding configuration in our experiments, reduces the encoder peak GPU memory from 6.314 GB to 4.382 GB and reduces the encoder extra peak memory by 39.14\%, while achieving a 1.41$\times$ speedup over non-chunked encoding. Shorter chunk sizes further reduce encoder memory usage, confirming that chunked encoding provides a practical memory-efficiency benefit.

Table~\ref{tab:encoder_param_efficiency} reports the encoder-side parameter footprint. The ``Encoder Params'' column denotes the total encoder-side parameter footprint after adding PEFT modules, while ``Encoder PEFT Params'' and ``PEFT Param. Ratio'' denote the PEFT-related parameters and their proportion, respectively.
Zipper-LoRA introduces additional shared and language-specific LoRA banks compared with 
Vanilla-LoRA, but remains more compact than Independent-LoRA. Specifically, Zipper-LoRA-Hard and Zipper-LoRA-Soft use 
173.53M encoder PEFT parameters, corresponding to 19.36\% of the encoder-side parameter footprint, 
whereas Independent-LoRA requires 283.13M encoder PEFT parameters and reaches 28.14\%. Therefore, 
Zipper-LoRA provides an intermediate design between Vanilla-LoRA and Independent-LoRA: it increases adaptation capacity to reduce inter-lingual interference while avoiding the 
larger parameter cost of maintaining a separate adapter for each language.

\subsection{Representation Visualization}
\label{subsec:repanalysis}

\begin{figure}[t]
  \centering
  \includegraphics[width=0.9\linewidth]{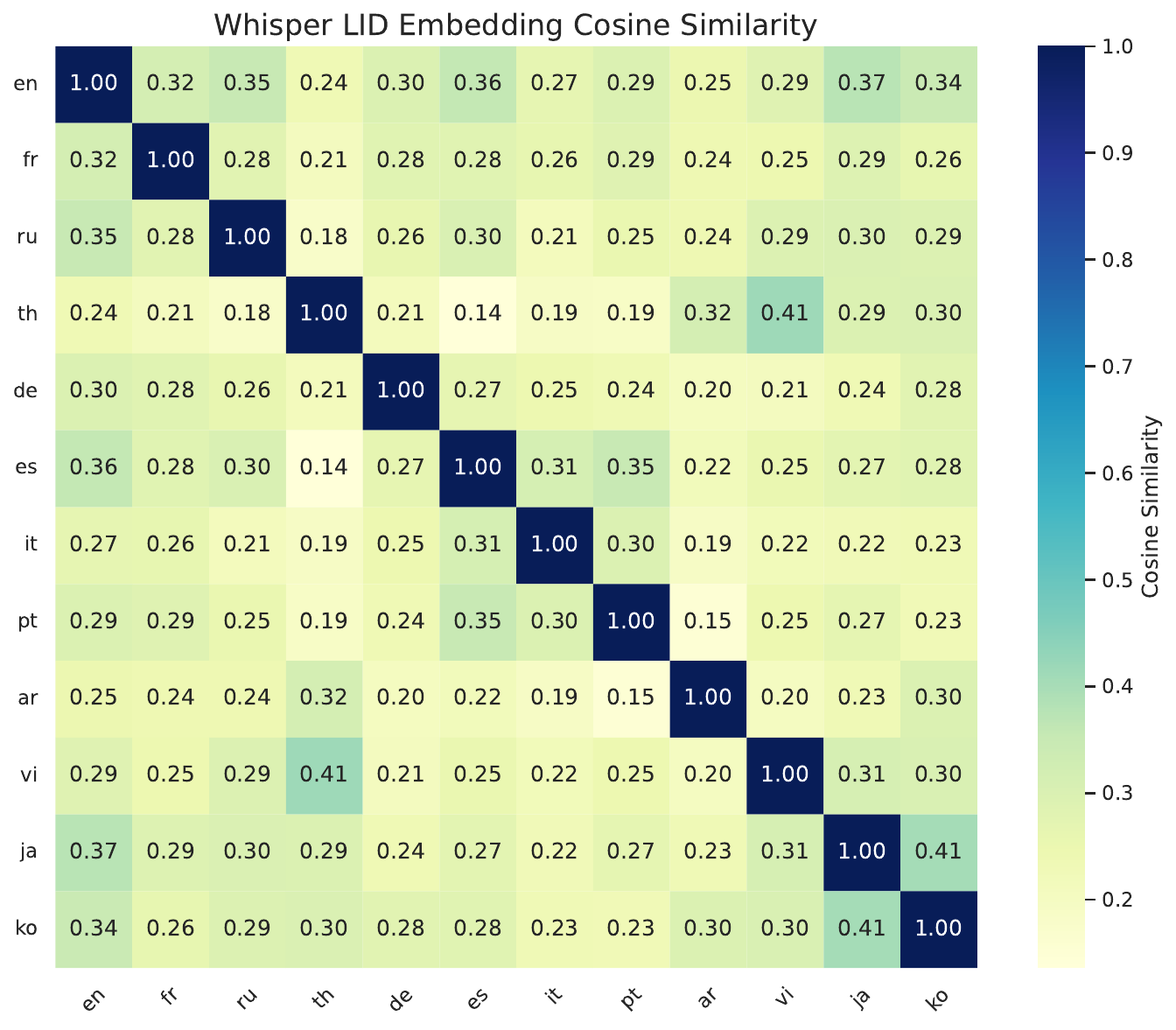}
  \caption{Cosine-similarity heatmap of Whisper language-ID (LID) embeddings for the 12 target languages.}
  \label{fig:lid_emb}
\end{figure}

We first analyze the language-ID (LID) embeddings provided by Whisper, which encode the model's 
prior notion of language similarity and are expected to reflect broad phonetic and acoustic similarities. 
Fig.\ref{fig:lid_emb} shows the pairwise cosine-similarity structure among the 12 target languages. 
We observe distinct high-similarity clusters between linguistically related languages; 
for example, Japanese (ja) and Korean ({ko}) exhibit a cosine similarity of 0.41, 
and Thai ({th}) and Vietnamese ({vi}) also reach 0.41. In contrast, more distant language 
pairs consistently show lower similarity values. This semantic topology indicates that 
Whisper's pretrained LID space already organizes languages into meaningful neighborhoods, 
motivating the use of LID embeddings as a routing anchor: nearby languages are more likely 
to benefit from knowledge transfer, whereas forcing distant languages to share adaptation capacity 
increases the risk of inter-lingual interference.

\begin{figure}[t]
  \centering
  \includegraphics[width=0.9\linewidth]{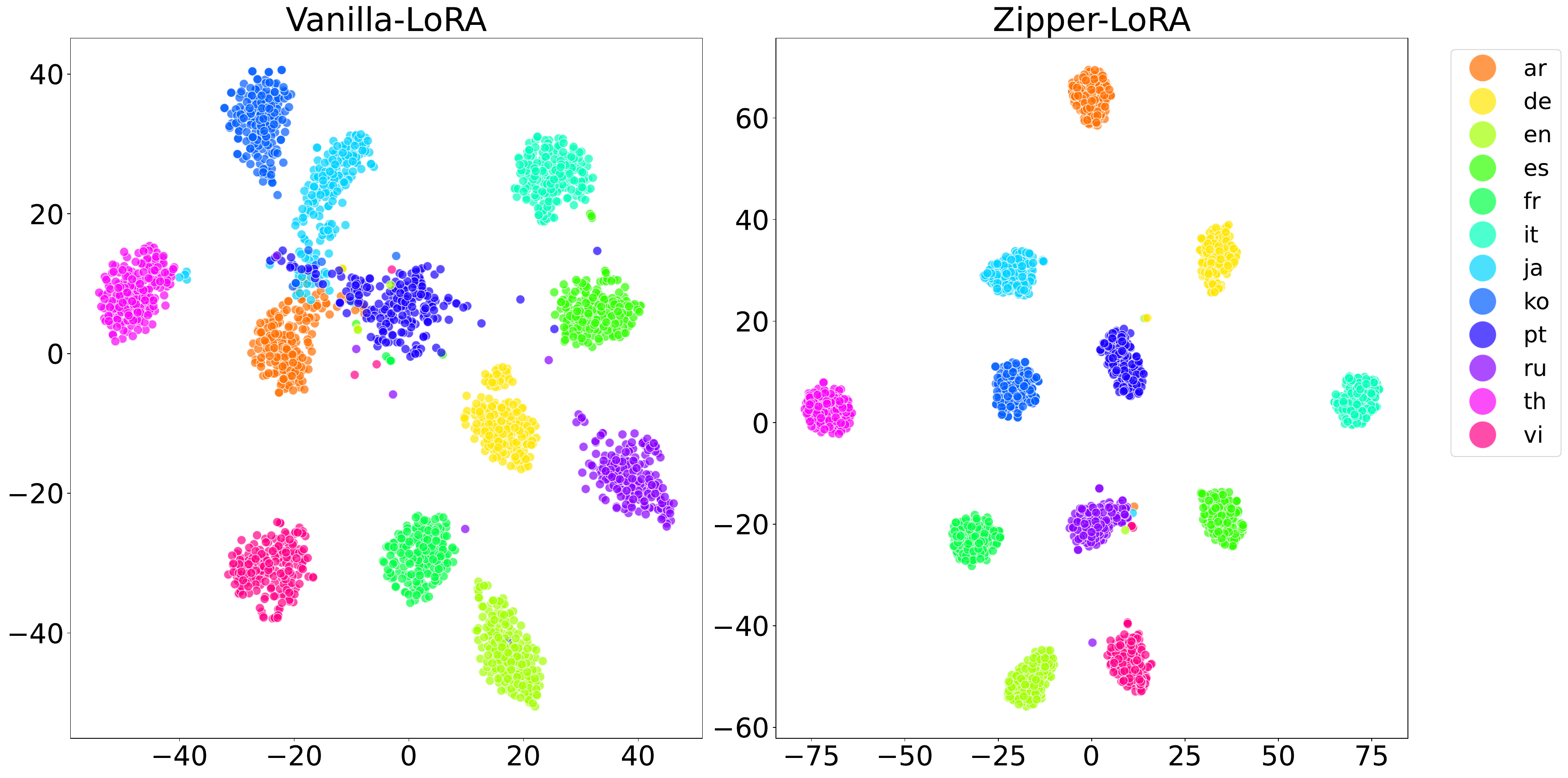}
  \caption{t-SNE visualization of encoder output representations across the 12 target languages.}
  \label{fig:tsne_enc}
\end{figure}

\begin{table}[t]
\centering
\caption{Per-language covariance determinant values computed from the 2-D t-SNE coordinates. Lower values indicate more compact language-specific clusters.}
\label{tab:covariance_determinant}
\resizebox{\columnwidth}{!}{%
\begin{tabular}{lccc}
\toprule
Language & Vanilla-LoRA & Zipper-LoRA-Soft & Relative Reduction \\
\midrule
ar & 180.19 & 245.13 & -36.04\% \\
de & 225.98 & 79.25 & 64.93\% \\
en & 209.74 & 68.11 & 67.53\% \\
es & 310.25 & 51.48 & 83.41\% \\
fr & 421.29 & 46.56 & 88.95\% \\
it & 90.13 & 182.22 & -102.17\% \\
ja & 627.35 & 252.78 & 59.71\% \\
ko & 301.17 & 55.35 & 81.62\% \\
pt & 1211.63 & 57.06 & 95.29\% \\
ru & 696.74 & 109.42 & 84.30\% \\
th & 77.51 & 43.17 & 44.30\% \\
vi & 360.05 & 118.93 & 66.97\% \\
\midrule
Average & 392.67 & 109.12 & 72.21\% \\
Median & 305.71 & 73.68 & 75.90\% \\
\bottomrule
\end{tabular}%
}
\end{table}

We then examine the encoder output representations after SFT ASR adaptation. Fig.~\ref{fig:tsne_enc} visualizes 
these representations using t-SNE, where each point corresponds to an utterance embedding and colors 
indicate languages. The qualitative evidence in Fig.~\ref{fig:tsne_enc} and the quantitative analysis in
Table~\ref{tab:covariance_determinant} jointly support four observations.
1) A clear contrast emerges between fully shared Vanilla-LoRA and Zipper-LoRA-Soft.
Under Vanilla-LoRA (left), multiple languages collapse into partially overlapping or densely entangled regions, 
indicating that the shared adaptation parameters induce cross-lingual interference; 
this effect is especially pronounced for low-resource languages, whose representations are more easily dominated 
and distorted by updates driven by higher-resource languages, leading to negative knowledge transfer. 
In contrast, Zipper-LoRA-Soft (right) yields more compact and well-separated language clusters, suggesting 
that it suppresses harmful inter-lingual interference while preserving language-specific factors. 
2) To further quantify this compactness, we compute the determinant of the covariance matrix for each
language-specific cluster based on the corresponding 2-D t-SNE coordinates. Since this determinant measures 
the generalized variance of a cluster, a smaller value indicates a more compact distribution in the embedding space. 
Table~\ref{tab:covariance_determinant} reports the detailed per-language covariance determinant values. Zipper-LoRA-Soft obtains lower covariance determinants than Vanilla-LoRA for 10 out of the 12 target languages. 
Moreover, the average determinant is reduced from 392.67 to 109.12, corresponding to a 72.21\% relative reduction, 
and the median determinant decreases from 305.71 to 73.68, corresponding to a 75.90\% relative reduction.
3) Although Arabic and Italian show larger determinants,
the overall trend and aggregate statistics consistently support the visual observation that Zipper-LoRA-Soft 
generally produces more compact encoder output representations. 4) This result provides qualitative and quantitative
evidence for the effectiveness of rank-wise composition: by enabling fine-grained sharing and specialization, 
Zipper-LoRA-Soft can exploit transferable structure where beneficial while preventing excessive coupling across 
unrelated or imbalanced languages, resulting in more stable multilingual adaptation.

\section{Conclusion}

Multilingual ASR adaptation under long-tailed data distributions must 
balance cross-lingual transfer and interference. Fully shared parameter-efficient tuning can
suffer from negative knowledge transfer, especially for under-represented languages, 
while fully language-specific tuning reduces inter-lingual interference but sacrifices beneficial 
sharing and can be severely constrained in the extremely low-resource conditions.

In this work, we introduced Zipper-LoRA, a rank-wise composition framework that decomposes 
adaptation capacity into shared and language-specific components and combines them through 
language-conditioned routing. This design enables fine-grained sharing where languages are 
compatible while preserving specialization to suppress harmful coupling. Across 12 target 
languages covering high and low-resource conditions, Zipper-LoRA variants consistently achieve 
strong performance and more balanced gains than shared baselines and fully decoupled alternatives. 
The improvements are particularly evident on long-tail languages, where controlled sharing 
provides additional benefit beyond complete parameter decoupling. We further showed that these 
advantages are robust across different encoder processing settings: although non-chunked encoding 
yields lower absolute error rates than chunked encoding due to the availability of broader context, 
the relative ranking and gains of Zipper-LoRA remain consistent. 
Finally, representation analyses support the proposed mechanism by revealing meaningful 
language structure in LID embeddings and clearer separation of adapted encoder representations 
under Zipper-LoRA, consistent with reduced inter-lingual interference.

Overall, our results suggest that rank-wise composition offers an effective 
and robust approach for practical multilingual adaptation, providing a better trade-off 
between cross-lingual transfer and inter-lingual interference than either fully shared or 
fully independent designs. Future work includes scaling to a larger and more diverse set of languages, 
exploring alternative routing signals beyond LID, and extending the approach to streaming or 
continual multilingual adaptation settings.

\section*{Acknowledgments}

This work was supported by the Natural Science Foundation of Shanghai
(Grant No. 25ZR1401277) and the National Natural Science Foundation of
China (Grant No. 62071302).

\bibliographystyle{IEEEtran}
\bibliography{refs}

\end{document}